\documentclass[11pt,a4paper]{article}
\usepackage{times,latexsym}
\usepackage{booktabs}
\usepackage{multirow}
\usepackage{multicol}
\usepackage{graphicx}
\usepackage{enumitem}
\usepackage{longtable}
\usepackage{float}
\usepackage{amssymb}
\usepackage{tabularray}
\usepackage{amsmath}
\usepackage{url}
\usepackage[T1]{fontenc}
\usepackage[table]{xcolor}
\usepackage[skins,breakable]{tcolorbox}
\usepackage{spverbatim}
\usepackage{algorithm}
\usepackage{algpseudocode}
\usepackage[acceptedWithA]{tacl2021v1}
\usepackage{xspace,mfirstuc,tabulary}

\newif\iftaclinstructions
\taclinstructionsfalse 
\iftaclinstructions
\renewcommand{\confidential}{}
\renewcommand{\anonsubtext}{(No author info supplied here, for consistency with
TACL-submission anonymization requirements)}
\newcommand{\instr}
\fi

\iftaclpubformat 

\else

\fi
\usepackage[T1]{fontenc}
\newcommand\thefont{\expandafter\string\the\font}

\title{GeLaCo: An Evolutionary Approach to Layer Compression}

\author{David Ponce$^{1,3}$ \quad Thierry Etchegoyhen$^{1}$ \quad Javier Del Ser$^{2,3}$ \\
        $^1$ Fundación Vicomtech, Basque Research and Technology Alliance (BRTA) \\
        $^2$ TECNALIA. Basque Research and Technology Alliance (BRTA) \\
        $^3$ University of the Basque Country UPV/EHU \\
        \texttt{\{adponce,tetchegoyhen\}@vicomtech.org} \\
        \texttt{javier.delser@tecnalia.com} 
        }
\date{}

\begin{document}
\maketitle

\begin{abstract}

Large Language Models have achieved remarkable performance across a large number of tasks, but face critical deployment and usage barriers due to substantial computational requirements. Model compression methods, which aim to reduce model size while preserving its capacity, are an important means to mitigate these issues. Promising approaches along these lines, such as structured pruning, 
typically require costly manual hyperparameter exploration or rely on local  heuristics that may run the risk of ignoring better solutions. 
In this work we introduce GeLaCo, an evolutionary approach to LLM compression via layer collapse. Our approach supports an efficient exploration of the compression solution space via population-based search and a novel layer collapse formulation based on parametrized weight merging, with a fitness function based on similarity over residual updates and language modeling KL divergence. GeLaCo also supports both single and multi-objective evolutionary compression search, establishing the first Pareto front estimation along compression and quality axes. We evaluate GeLaCo solutions via both perplexity-based and generative evaluations over foundational and instruction-tuned models, outperforming state-of-the-art alternatives.
\end{abstract}

\section{Introduction}

The remarkable success of Large Language Models (LLM) across diverse natural language processing tasks \cite{radford2019language,brown2020language,chang2024survey} has come at the cost of substantial computational requirements. Modern LLMs, ranging from billions to trillions of parameters, demand considerable memory and processing power for both training and inference, creating barriers to their widespread deployment and usage. This computational burden is particularly challenging for edge devices, mobile applications, and resource-constrained environments, where reduced memory footprint and efficient inference are essential for practical adoption.

Model compression has emerged as an important field of research to address these challenges \cite{xu2023survey,zhu2024survey}. Traditional compression techniques mainly include quantization, reducing numerical precision \cite{gholami2022survey,jin2024comprehensive}; knowledge distillation, which transfers capabilities to smaller models \cite{gou2021knowledge,xu2024kdsurvey}; and pruning, where redundant or less impactful parameters are removed \cite{wang2020structured,cheng2024survey}. Among the latter, recent work by \citet{yang-etal-2024-laco} introduced the idea of merging consecutive layers into a combined representation, partially preserving layer information rather than directly pruning parts of the network, an approach we built upon in this work. While compression approaches have demonstrated their potential, they often require costly empirical work to determine optimal model variants, with the added risk of leaving potentially better compression solutions unexplored.

Evolutionary Algorithms (EA) can provide a more principled exploration of the compression solution space, generating and filtering different model configurations via appropriate fitness functions. Recent work has explored LLM compression under evolutionary approaches. For instance, \citet{sieberling2024evopress} tackle non-uniform unstructured pruning, quantization and layer pruning via EA, using Kullback-Leibler (KL) divergence as a fitness function. Building upon this approach, \citet{tang2025darwinlm} address structured pruning at the component level by including limited training steps as part of the selection process. Alternatively, \citet{huang2025towards} explored an EA approach for LLM self-pruning, with the evolutionary search performed by the model itself over its own internal representations. To our knowledge, evolutionary compression via layer collapse has not yet been explored.

In this work we introduce GeLaCo, an efficient evolutionary approach to search for optimized LLM compression solutions, based on a novel formulation of layer collapse centered on parametrized weight difference merging, with a fitness function that preserves cumulative residual contributions while minimizing KL divergence over the final language modeling state. GeLaCo can formulate compression as either a single-objective problem, with a fixed compression target, or as a multi-objective optimization problem that simultaneously considers compression ratio and model performance. Via the latter, we uncover Pareto-optimal solutions that reveal current trade-offs between efficiency and capability.

We conduct our experiments across different model sizes and types, including base and instruction-tuned variants, and evaluate both perplexity-based and generative performance across a wide range of benchmarks. Our results indicate that GeLaCo can  lead to compression solutions that outperform alternative compression approaches. Furthermore, we demonstrate that GeLaCo solutions for the bi-objective compression problem are Pareto-dominant overall, while also revealing critical compression thresholds beyond which model performance degrades significantly. To support reproducibility and future research, our code and  models will be made publicly available upon publication.

\section{Related Work}



\begin{figure*}[t!]
    \centering
    \resizebox{2\columnwidth}{!}{
        \includegraphics[width=2\columnwidth]{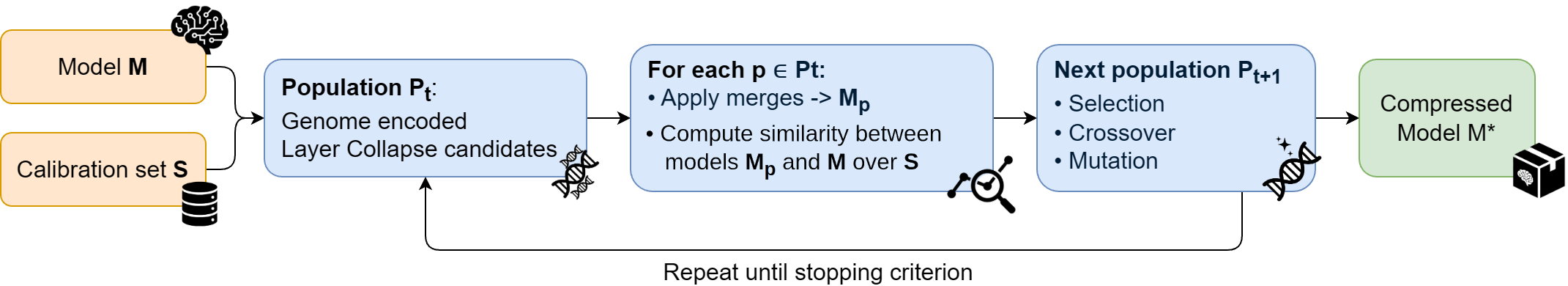}}
    \caption{Overview of GeLaCo. Given a model and calibration set, layer-merged candidates are evolved via selection, crossover, and mutation, using a fitness function that measures residual similarity and KL divergence. This process iterates until convergence, yielding a compressed model that aims to minimize degradation.}
    \label{fig:overview}
\end{figure*}

\subsection{Standard Compression}

Several standard techniques have been proposed to reduce the size of LLMs in particular  \cite{xu2023survey,zhu2024survey}. For instance, quantization is an effective compression technique where numerical precision is reduced for model weights and/or activations, either during training or as a post-training optimization process \cite{whittaker2001quantization,wang2023bitnet,xiao2023smoothquant,lin2024awq,huang2024billm,ma2024era,xu2024onebit,liu2024spinquant,liu2024llm,dettmers2023spqr}. Knowledge distillation is another standard approach to model compression, via the transfer of knowledge from large models to smaller predefined models \cite{yang2024survey,stanton2021does,gu2023minillm}. This method can be combined with either quantization techniques \cite{kim2019qkd,liu2023llm} or model pruning \cite{muralidharan2024compact,kim2021pqk}, to achieve higher compression rates while preserving model quality.

Our work is mainly related to model pruning, where compression targets the removal of model parameters. This is a typical approach to compressing Transformer models \cite{vaswani2017attention}, the prevalent architecture for LLMs. Among these methods, unstructured pruning typically addresses parameter sparsity. In this line, \citet{frantar2023sparsegpt} proposed the SparseGPT approach, where LLMs undergo one-shot pruning, with minimal perplexity increase, via a sparse regression solver. Alternatively, the SliceGPT approach \cite{ashkboos2024slicegpt} relies on a post-training method where sparse weight matrices are replaced with smaller dense matrices. Other approaches rely on shared parameters, for instance \citet{cao2024head}, who exploit shareable attention parameters either directly or via post-training.

Structured pruning approaches, where specific layers of the model are removed, are in line with our GeLaCo approach. In the ShearedLlama approach \cite{xia2023sheared}, Llama-2 models \cite{touvron2023llama} are compressed via both targeted structured pruning and dynamic batch loading. \citet{men2024shortgpt} proposed ShortGPT, where redundant Transformer blocks are eliminated based on Block Information scores, derived from block input and output similarity. \citet{ma2023llm} describe LLM-Pruner, a method that relies on gradient information to identify components of the model which can be pruned without affecting the overall capacity of the model. Whereas most structured pruning approaches identify and remove selected parts of the model, \citet{yang-etal-2024-laco} introduced LaCo, where layers are merged depending on layer-wise cosine similarity differences, measuring model deviation via small calibration data and discarding merge operations that fall under a predefined similarity threshold. They showed that the best LaCo configuration, empirically determined, outperformed state-of-the-art approaches such as LLM-Pruner and SliceGPT. More recently, related structured compression methods have explored combining or merging layers to better preserve model capacity, as in GPTailor \cite{su2026gptailor} and CoMe \cite{wang2025layer} as well as search-based optimization of non-uniform sparsity across layers, as in Týr-the-Pruner \cite{li2025trthepruner}.

Despite their merits, most current approaches require costly empirical evaluations of different compression schemes or rely on local heuristics, at the risk of ignoring better compression solutions. We address these limitations by integrating layer merging into an efficient evolutionary framework, showing that this can lead to better compression solutions in most scenarios. We also propose a novel parametrized weight merging approach and a novel similarity objective that aims to preserve cumulative residual contributions, both improving over the main state-of-the-art alternatives.

\subsection{Evolutionary Compression}

Recently, EAs have gained increasing attention as a promising approach to optimize the compression of large foundation models. A notable advancement in this domain is EvoPress \citep{sieberling2024evopress}, which formulates model compression as a general optimization problem and employs EAs to discover compression profiles for LLMs. This approach enables dynamic, non-uniform compression by adaptively selecting compression levels at the block or layer level, thereby facilitating fine-grained optimization that preserves model capabilities while achieving substantial size reductions across diverse compression techniques.

Another line of work explores self-pruning mechanisms, wherein LLMs autonomously perform the evolutionary search process, leveraging their internal representations of redundancy to generate, evaluate, and optimize pruning configurations with minimal human intervention \citep{huang2025towards}. A further contribution in this space is DarwinLM \citep{tang2025darwinlm}, which introduces a training-aware structured pruning strategy. DarwinLM integrates a multistage training process within the evolutionary framework, progressively increasing token usage and eliminating under-performing models at each selection step. Finally, \citet{wu2025evop} present EvoP, an alternative EA compression approach based on data calibration and layer pruning that outperformed heuristic structure pruning alternatives.

Our work follows an EA approach to model compression, but differs from the aforementioned studies in important ways. First,  we explore the first EA approach based on layer collapse, thereby extending the family of EA methods for model compression. Secondly, we address both single and multi-objective optimization, revealing the first Pareto front approximations in terms of compression ratio and model quality. Finally, we provide novel results in terms of post-training EA-compressed models for both base and instruction-tuned models, while also providing the first generative evaluation results for the latter.
\section{GeLaCo}

Our approach to model compression adopts an evolutionary framework, wherein a population of candidate compression solutions evolves according to either a single compression objective, or multiple objectives such as compression ratio and model quality. In what follows, we describe our approach to layer compression and the core components of the evolutionary framework. 

\subsection{Layer Collapse}

Our approach builds upon the idea of layer collapse, whereby the weights of each layer included in a given structured compression operation are merged into the final compressed layer via weight difference updates. The original idea for this method was proposed by \citet{yang-etal-2024-laco}, referred to as LaCo. It should be noted however that there is a subtle but important discrepancy between the theoretical LaCo approach and its official implementation\footnote{\url{https://github.com/yangyifei729/LaCo}}. Whereas the latter corresponds to their published results, per our experiments, it also differs from the published algorithm in that the implementation eventually leads to merging only the weights of the final layer in the compressed sequence (see Appendix~\ref{app:laco_implementation}); 
in our experiments using the weight merging definition and hyper-parameters reported in the original approach yielded no effective compression, as all candidate configurations fell under the cosine similarity threshold.
Taking these results into account, we designed a novel approach to layer collapse, which outperforms both the original layer merging and layer dropping variants.

Our layer collapse variant is based on parametrized weight merges, meant to provide the evolutionary search with a solution space where the contribution of compressed layers into the final weights can vary individually across solutions. To provide additional flexibility, we further consider separately the merging of the parametrized weights for the self-attention and feed-forward layers.

Thus, given a model with layers $\{L_0,\ldots,L_{L-1}\}$, and denoting as $\Theta^{\mathrm{att}}_k$ and $\Theta^{\mathrm{ffn}}_k$ the attention and feed-forward parameters of layer $k$, respectively, we define the parameters of the merged layer corresponding to the collapse of a contiguous span $[i,\ldots,j]$ as:

{\small
\begin{equation}
\begin{aligned}
\Theta^{\mathrm{att}}_{i:j}
&=
\Theta^{\mathrm{att}}_{i}
+
\sum_{k=i+1}^{j}
\lambda^{\mathrm{att}}_{k}
\left(
\Theta^{\mathrm{att}}_{k}-\Theta^{\mathrm{att}}_{i}
\right) \\
\Theta^{\mathrm{ffn}}_{i:j}
&=
\Theta^{\mathrm{ffn}}_{i}
+
\sum_{k=i+1}^{j}
\lambda^{\mathrm{ffn}}_{k}
\left(
\Theta^{\mathrm{ffn}}_{k}-\Theta^{\mathrm{ffn}}_{i}
\right)
\end{aligned}
\label{eq:module_merge}
\end{equation}}

where $\lambda^{\mathrm{att}}_{k}$ and $\lambda^{\mathrm{ffn}}_{k}$ control the contribution of layer $k$. This formulation preserves the base parametrization, while allowing the merged block to interpolate along the directions induced by the removed layers.

\subsection{Functional Preservation}

Compressing a contiguous span of Transformer layers into a single surrogate layer requires preserving the effect of the removed layers on the hidden state. In modern decoder-only Transformers \cite{grattafiori2024llama, qwen3technicalreport}, each layer contributes to this effect through two sequential updates to the residual stream, first via self-attention and then via the feed-forward module. Therefore, the functional preservation of a span of layers requires preserving the residual contributions introduced across these successive updates.

Formally, for a given layer $k$, we write the induced transformation as in Equation~\eqref{eq:layer_transform}, where $h$ denotes the input vector, $Attn_{k}(h)$ the self-attention layer transformation, and $FFN_{k}(h + Attn_{k})$ the feed-forward transformation: 

{
\small
\begin{equation}
\mathcal{L}_k(h)
=
h
+ \operatorname{Attn}_k(h)
+ \operatorname{FFN}_k\!\big(h + \operatorname{Attn}_k(h)\big)
\label{eq:layer_transform}
\end{equation}
}

For a contiguous span of layers $L_\ell,\ldots,L_{\ell+m}$ the cumulative transformation is therefore:

{
\small
\begin{equation}
h_{\ell+m+1}
=
(\mathcal{L}_{\ell+m}\circ\cdots\circ\mathcal{L}_\ell)(h_\ell)
\label{eq:layer_composition}
\end{equation}
}

When collapsing this span into a single surrogate layer $L_{\ell:\ell+m}$, the merged computation is reduced to two sequential residual updates:

{
\small
\begin{equation}
\begin{aligned}
\mathcal{L}_{\ell:\ell+m}(h)
&= h + \operatorname{Attn}_{\ell:\ell+m}(h) \\
&\quad + \operatorname{FFN}_{\ell:\ell+m}\!\bigl(
h + \operatorname{Attn}_{\ell:\ell+m}(h)
\bigr)
\end{aligned}
\label{eq:merged_layer_transform}
\end{equation}
}

Since the residual stream is not normalized after each update, the magnitude of residual contributions grows with depth in Pre-Norm architectures \cite{li2026siamesenorm}. Matching directions alone is therefore insufficient to account for the contributions of the removed layers. We thus introduce independent scaling factors for the attention and feed-forward output projections:

{
\small
\begin{equation}
\widetilde{W}^{O}_{i:j}=\alpha \, W^{O}_{i:j},
\quad
\widetilde{W}^{D}_{i:j}=\beta \, W^{D}_{i:j},
\label{eq:gated_outputs}
\end{equation}
}

where $W^{O}_{i:j}$ denotes the attention output projection and $W^{D}_{i:j}$ the feed-forward down projection of the merged layer, while $\alpha$ and $\beta$ control the effective contribution of each update to the residual stream.

\subsection{Evolutionary Framework}

Identifying effective combinations of merge operations is challenging, as the search space grows exponentially with the number of possible merge configurations. Moreover, unlike greedy or threshold-based strategies, where decisions are made locally, the effectiveness of any single merge depends on its interaction with the rest of the compression scheme. Local heuristics are therefore insufficient to discover globally effective compression solutions. To address this, we adopt a population-based evolutionary algorithm that maintains a population of candidate solutions, where each individual represents a specific configuration of layer-merge operations.

\subsubsection{Fitness Function}
\label{sec:fitness}

Our fitness function is designed to preserve the cumulative residual contributions of the merged layers. To evaluate each candidate solution, we perform inference on a small calibration set of sentences. For every layer in the merged model, we compare the attention and feed-forward residual contributions of the compressed layer against the sum of the corresponding residual contributions over the original span. The fitness function thus measures how well the compressed layer reproduces the aggregated residual effect of the removed layers. Similarity is computed using a score that jointly captures vector direction and magnitude. For two vectors $v$ and $u$, we thus compute similarity as:
\begin{equation}
S(u,v)=\frac{2\langle u,v\rangle}{\|u\|^2+\|v\|^2}
\label{eq:fitness_similarity}
\end{equation}

To also preserve the underlying language modeling behavior, we compare the output distributions of the original and compressed models through the token-level Kullback-Leibler (KL) divergence. Since lower KL indicates better agreement, we convert it into a bounded maximization score through $e^{-KL}$, so that higher values correspond to better solutions.

Let $S_{\mathrm{att}}$ and $S_{\mathrm{ffn}}$ denote the average similarity scores obtained from this comparison over all merged layers and calibration sentences for the attention and feed-forward residual branches, respectively. The final fitness value is defined as:
\begin{equation}
\mathcal{F}
=
\frac{1}{3}
\left(
S_{\mathrm{att}}
+
S_{\mathrm{ffn}}
+
e^{-KL}
\right)
\label{eq:fitness_function}
\end{equation}

This yields an objective that encourages compressed layers to reproduce the aggregated residual behavior of the removed span while remaining close to the original model at the token distribution level.

\subsubsection{Candidate Solutions}

To enable evolutionary search, candidate solutions must be encoded as genomes that can be modified through crossover and mutation and later decoded into valid sequences of layer-merge operations.

\paragraph{Solution Encoding.} For a model with $L$ layers and a target compression depth $T$, let $R = L - T$ denote the number of layers to remove. Since each non-degenerate merge over a contiguous span removes at least one layer, any compression requires at most $R$ merge operations. Each candidate solution is then encoded as a mixed integer-real genome. 

The integer part assigns a priority $t_b \in \{0,\dots,R\}$ to each boundary $b$ between consecutive layers which determine the order in which boundaries become active during decoding. The real-valued part stores the parameters required to instantiate the merged layers. First, each original layer \(i\in\{0,\dots,L-1\}\) is associated with one attention coefficient and one feed-forward coefficient,
\[
\small
\boldsymbol{\lambda}^{\mathrm{att}}=
(\lambda^{\mathrm{att}}_0,\dots,\lambda^{\mathrm{att}}_{L-1}) \hspace{0.3cm}
\boldsymbol{\lambda}^{\mathrm{ffn}}=
(\lambda^{\mathrm{ffn}}_0,\dots,\lambda^{\mathrm{ffn}}_{L-1})
\]
which determine how much the attention and feed-forward modules of each layer contribute when that layer participates in the merge operation defined in Eq.~\eqref{eq:module_merge}. In addition, each merge step \(r\) has associated values \(g^{\mathrm{att}}_r\) and \(g^{\mathrm{ffn}}_r\), to control the residual scaling applied to the attention and feed-forward modules respectively, as defined in Eq.~\ref{eq:gated_outputs}. It also has associated values \(
\hat{\lambda}^{\mathrm{att}}_r, \hat{\lambda}^{\mathrm{ffn}}_r
\), which are assigned to the surrogate layer created by that merge step and determine its contribution if it later participates in any subsequent merge operation. Hence, the full encoding contains \((L-1)+2L+4R\) variables in total per solution.

\paragraph{Solution Decoding.} Decoding maps the full candidate encoding into a sequential list of merge operations. Starting from the original layer sequence, boundaries are activated in decreasing priority order based on $t_b$, with ties activating simultaneously. These priorities determine which contiguous groups of layers are collapsed at each stage. When a merge is applied, the operation collects the coefficients \(\lambda^{\mathrm{att}}\) and \(\lambda^{\mathrm{ffn}}\) associated with the layers being merged together with residual gates \(g^{\mathrm{att}}\) and \(g^{\mathrm{ffn}}\). Decoding proceeds until the target depth $T$ is reached, yielding  the ordered list of merge operations. A detailed algorithmic description is provided in Appendix~\ref{app:decoding_algorithm}.

\subsubsection{Search Objectives}

For single-objective optimization, GeLaCo employs a standard genetic algorithm with mixed integer-real encoding operators, where solutions are sorted and selected based on our module-wise similarity metric. For multi-objective optimization, GeLaCo relies on the Non-dominated Sorting Genetic Algorithm (NSGA-II) to maintain population diversity. The evolutionary process terminates when a predetermined maximum number of fitness evaluations is reached. Further implementation details are provided in Appendix~\ref{app:cache}.

\paragraph{Single Objective.} For single-objective optimization, we formulate the problem as similarity maximization subject to exact compression objectives. In our formulation, compression ratio represents the proportion of layers removed from the original model through merging operations. A compression ratio of 0.25 indicates that 25\% of the original layers have been collapsed, yielding a model with 75\% of the original layer count.

To ensure that solutions satisfy the target compression after crossover and mutation operations, we employ a repair mechanism similar to \citet{wu2025evop}. The repair process systematically toggles, modifies, or creates merge operations until solutions achieve the exact target compression ratio, maintaining search focus on feasible configurations. Solutions that deviate from this target after a predetermined number of trials are penalized with a worst possible fitness score (-1.0), to avoid exploring these configurations.

\paragraph{Multiple Objectives.} For multi-objective optimization, we consider the joint optimization of compression ratio and similarity preservation, without predefined compression targets. To this end, the candidate encoding includes an additional real-valued decision variable in the range $[0,1]$, which is free to evolve and determines the target compression ratio associated with each solution. The retention of solutions based on non-dominated sorting and crowding distance enables the discovery of a Pareto front approximation that captures the trade-offs between model similarity and compression ratio.

\section{Experimental Setup}

Our experiments are designed to assess several aspects of GeLaCo solutions, and gain further understanding of its current strengths and limitations. To this effect, we first evaluate different fitness functions, then addressing single-objective approach comparisons, multi-objective Pareto front assessment, and evaluations of post-training base and instruction models.

\paragraph{Operational Framework.}
Evolutionary search was performed with the reference JMetalPy framework \cite{benitez2019jmetalpy}. Since our solutions are defined over a mixed integer-real search space, we used a composite SBX crossover combining integer-adapted and real-valued SBX operators, both with a probability of $0.9$ and a distribution index of $20.0$. The mutation strategy analogously relies on composite polynomial mutation over integer and real-valued variables, with probability scaled to the full problem dimensionality $(1.0 / N)$, where $N$ is the length of the genome, and a distribution index of $20$. For single-objective experiments, we employed a population of 64 individuals over 4,096 fitness evaluations. For multi-objective Pareto optimization, we used a larger population of 128 individuals over 16,192 evaluations. Following prior evidence that the final Transformer layer is particularly sensitive to pruning \citep{gromov2025the}, we kept it fixed in our implementation and excluded it from the set of admissible merge operations. We report average results for GeLaCo over three independent runs for all compression experiments, indicating standard deviations in all cases.

\paragraph{Language Models.}

We evaluated our compression framework using Llama variants as foundational base models: Llama-2 7B and 13B models~\citep{touvron2023llama}; Llama-3.1 8B and 70B~\citep{grattafiori2024llama}. We also included Llama-3.1 8B Instruct to investigate how evolutionary compression affects instruction-following capabilities, providing insights into the preservation of task-specific behaviors beyond general language modeling performance. Additionally, we evaluated GeLaCo on the Qwen3 8B~\citep{qwen3technicalreport} model to assess the portability of GeLaCo to Transformer architectures beyond those within the Llama family.

\paragraph{Baselines.}

We followed a similar experimental setup to LaCo, evaluating against the same structured compression baselines: LLM-Pruner \citep{ma2023llm} and SliceGPT \citep{ashkboos2024slicegpt}, while also including LaCo itself using their reported hyper-parameters and their official implementation (equivalent to layer dropping, as previously discussed). To ensure a fair evaluation, we replicated and ran our experiments for all baseline methods. We note that the current implementation of SliceGPT\footnote{\url{https://github.com/microsoft/TransformerCompression}.} does not support the Llama 3.1 family of models; thus, we could only provide results for Llama-2 models for this method. We also included EvoPress as an evolutionary alternative, using its structured pruning variant based on block or layer dropping, as this setting is more directly comparable to our layer collapse approach. We also report results for DarwinLM \citep{tang2025darwinlm}, using their released post-trained model\footnote{\url{https://huggingface.co/Shengkun/DarwinLM-4.6B}.}, considering the computational cost of post-training. Neither SliceGPT nor LLM-Pruner currently support Qwen3 models and we therefore excluded these approaches from our evaluation on this family of models.

\paragraph{Post-training.}
Since compression methods typically incur performance degradation, post-training is a standard step to stabilize the network and at least partially recover model capability. In Section~\ref{sec:post-training} we evaluate post-training over the compression configurations discovered by GeLaCo for the foundational Llama-3.1 8B and the instruction-tuned Llama-3.1 8B Instruct models. We use the hyper-parameters described in Appendix~\ref{app:hyperparms}.

\paragraph{Datasets.}
To compute similarity fitness during the evolutionary optimization phase, we used 64 sentences randomly sampled from the English Wikipedia dataset\footnote{\url{https://huggingface.co/datasets/wikimedia/wikipedia}}; to ensure a fair comparison, we employ the same 64-sentence subset when replicating LaCo and EvoPress baseline results. For the instruction models, we selected the LaMini Instruction dataset \citep{wu2023lamini-lm}. For post-training, we leveraged 10B tokens from the Fineweb-Edu dataset \citep{lozhkov2024fineweb-edu}.

\paragraph{Benchmarks.} We evaluate compressed model performance across multiple task-agnostic benchmarks using the \textit{lm-evaluation-harness} framework~\citep{eval-harness} for zero-shot evaluation on standard benchmarks: BoolQ~\citep{clark-etal-2019-boolq}, PIQA~\citep{bisk2020piqa}, HellaSwag~\citep{zellers2019hellaswag}, WinoGrande~\citep{sakaguchi2021winogrande}, ARC-easy and ARC-challenge~\citep{clark2018think}, OpenbookQA~\citep{mihaylov-etal-2018-suit}, LogiQA~\citep{liu2020logiqa}, SciQ~\citep{welbl2017crowdsourcing}, and MMLU \citep{hendrycks2020measuring}. We report normalized accuracy where available, otherwise standard accuracy.

To assess the instruction-following capabilities of instruction-tuned models, we evaluate their performance on the Just-Eval dataset~\citep{Lin2023ReAlign}, a suite of evaluation benchmarks, using the Prometheus 2 model~\citep{kim-etal-2024-prometheus} as a judge\footnote{Evaluation rubric provided in Appendix~\ref{app:eval-rubric}.}. Specifically, we use the first 800 samples from the \textit{just-eval-instruct} dataset, which focus on helpfulness and general instruction-following capabilities, excluding the 200 safety-focused examples to maintain our evaluation scope on core instruction-following performance.


\paragraph{Computational Setup.}
Single-objective search was performed on a single NVIDIA A40 with 48GB of VRAM. The same setup was used for multi-objective search, except for Llama-3.1 70B, which required two NVIDIA H100s with 80GB each. Model evaluation was conducted on a single A40 for all models except Llama-3.1 70B, which was evaluated on a single H100. Post-training was carried out on four H100s. A detailed throughput comparison between GeLaCo and competing pruning methods is provided in Appendix~\ref{app:throughput}.

\section{Results}

\subsection{Fitness Function Evaluation}
\label{sec:fitness-function-eval}
To evaluate our proposed fitness function, we contrasted four optimization objectives that capture different aspects of compressed models: KL divergence, which measures similarity between the output distributions of the original and compressed models and is used by evolutionary pruning methods such as DarwinLM or EvoPress; perplexity, which reflects language modeling performance; our proposed Residual Similarity (RS) objective, which measures preservation of cumulative residual contributions; and a combined KL+RS objective. For these experiments, we applied GeLaCo to Llama-2 7B at a medium compression ratio of 0.281. The results are presented in Table~\ref{tab:fitness_comparison}.

\begin{table}[h]
\centering
\footnotesize
\resizebox{1\columnwidth}{!}{\begin{tabular}{lcccc}
\toprule
& \multicolumn{4}{c}{\textbf{Optimization Objective}} \\
\cmidrule(lr){2-5}
			& \textbf{PPL}			& \textbf{KL}			& \textbf{RS}  & \textbf{KL+RS}\\
\midrule
\textbf{Evaluation Metric} \\
\midrule
KL Divergence ↓			& 0.643			& \textbf{0.618} 		& 1.624				& 1.604  \\
Perplexity ↓			& \textbf{15.212}		& 15.542 		& 45.294	 		& 41.541  \\
\multicolumn{5}{l}{\textit{Residual Similarity} ↑} \\
\quad Attention	& 0.779		& 0.769		& \textbf{0.953} 	& 0.932 \\
\quad FFN		& 0.748		& 0.747		& \textbf{0.945} 	& 0.940 \\
\midrule
\multicolumn{5}{l}{\textbf{Task Performance}} \\
\midrule
BoolQ			& 0.577			& 0.597			& 0.622		& \cellcolor{teal!20}\textbf{0.623} \\
PIQA			& \cellcolor{teal!20}\textbf{0.721}			& 0.696			& 0.655		& 0.675 \\
HellaSwag		& \cellcolor{teal!20}\textbf{0.559}			& 0.553			& 0.507		& 0.538 \\
WinoGrande		& 0.543			& 0.527			& 0.608		& \cellcolor{teal!20}\textbf{0.634} \\
ARC-easy		& 0.537			& \cellcolor{teal!20}\textbf{0.541}			& 0.487		& 0.540 \\
ARC-challenge	& 0.306			& 0.314			& 0.330		& \cellcolor{teal!20}\textbf{0.340} \\
OpenBookQA		& 0.344			& 0.338			& 0.340		& \cellcolor{teal!20}\textbf{0.366} \\
MMLU			& 0.247			& 0.260			& 0.262		& \cellcolor{teal!20}\textbf{0.376} \\
Average			& 0.479			& 0.478			& 0.476		& \cellcolor{teal!20}\textbf{0.512} \\
\bottomrule
\end{tabular}}
\caption{Impact of fitness function choice with GeLaCo on Llama-2 7B at 0.281 compression:  KL divergence (KL), perplexity (PPL), our proposed residual similarity (RS), and the combination of KL and RS (KL+RS).}
\label{tab:fitness_comparison}
\end{table}

As expected, each variant achieved the best performance on its own target metric, validating the effectiveness of the evolutionary search. On downstream tasks, however, while individual metrics remained competitive, with the highest score on some tasks, the best overall performance was achieved by the combined KL+RS objective, with the highest average task performance and the highest scores on the majority of tasks. Considering these results, all experiments reported in what follows were performed with the KL+RS similarity metric as our GeLaCo fitness function.

\subsection{Single-objective Optimization}
\label{sec:single-objective}
\begin{table*}[t]
\footnotesize
\centering
\resizebox{\textwidth}{!}{\begin{tabular}{ccccccccccccc}
\toprule
\textbf{Model} & \textbf{Ratio} & \textbf{Method} & \textbf{boolq} & \textbf{piqa} & \textbf{hellaswag} & \textbf{winogrande} & \textbf{arc\_easy} & \textbf{arc\_challenge} & \textbf{openbookqa} & \textbf{mmlu} & & \textbf{average} \\
\cmidrule{1-11} \cmidrule{13-13}
\multirow{18}{*}{\rotatebox[origin=c]{90}{Llama-2-7b}} 
& 0.000 & Dense & 0.778 & 0.791 & 0.760 & 0.691 & 0.745 & 0.462 & 0.442 & 0.418 & & 0.636\\
\cmidrule{2-11} \cmidrule{13-13}
& \multirow{5}{*}{0.125} & LaCo & 0.558 & 0.751 & 0.707 & \cellcolor{teal!20}\textbf{0.688} & 0.666 & 0.416 & 0.386 & \cellcolor{teal!20}\textbf{0.394} & & 0.571\\
& & LLM-Pruner & 0.681 & \cellcolor{teal!20}\textbf{0.779} & \cellcolor{teal!20}\textbf{0.719} & 0.671 & \cellcolor{teal!20}\textbf{0.676} & \cellcolor{teal!20}\textbf{0.422} & 0.422 & 0.271 & & \cellcolor{teal!20}\textbf{0.580}\\
& & SliceGPT & 0.581 & 0.737 & 0.666 & 0.666 & 0.630 & 0.392 & \cellcolor{teal!20}\textbf{0.430} & 0.300 & & 0.550\\
& & EvoPress & \cellcolor{teal!20}\textbf{0.689} & 0.773 & 0.696 & 0.623 & 0.644 & 0.392 & 0.392 & 0.300 & & 0.564 \\
& & GeLaCo & 0.655 $\pm$ 0.029 & 0.757 $\pm$ 0.002 & 0.706 $\pm$ 0.002 & \cellcolor{teal!20}\textbf{0.682 $\pm$ 0.007} & \cellcolor{teal!20}\textbf{0.680 $\pm$ 0.006} & 0.418 $\pm$ 0.002 & 0.400 $\pm$ 0.005 & 0.351 $\pm$ 0.027 & & \cellcolor{teal!20}\textbf{0.581 $\pm$ 0.005}\\
\cmidrule{2-11} \cmidrule{13-13}
& \multirow{5}{*}{0.281} & LaCo & \cellcolor{teal!20}\textbf{0.622} & 0.705 & 0.572 & 0.579 & 0.516 & \cellcolor{teal!20}\textbf{0.344} & 0.358 & 0.237 & & 0.491\\
& & LLM-Pruner & 0.497 & \cellcolor{teal!20}\textbf{0.740} & \cellcolor{teal!20}\textbf{0.606} & 0.568 & \cellcolor{teal!20}\textbf{0.560} & 0.333 & \cellcolor{teal!20}\textbf{0.378} & 0.231 & & 0.489\\
& & SliceGPT & 0.429 & 0.656 & 0.520 & 0.624 & 0.543 & 0.328 & 0.346 & 0.233 & & 0.460\\
& & EvoPress & 0.524 & 0.721 & 0.543 & 0.532 & 0.529 & 0.301 & 0.340 & 0.232 & & 0.465 \\
& & GeLaCo & \cellcolor{teal!20}\textbf{0.623 $\pm$ 0.003} & 0.679 $\pm$ 0.003 & 0.546 $\pm$ 0.013 & \cellcolor{teal!20}\textbf{0.642 $\pm$ 0.009} & 0.532 $\pm$ 0.006 & \cellcolor{teal!20}\textbf{0.344 $\pm$ 0.005} & 0.358 $\pm$ 0.006 & \cellcolor{teal!20}\textbf{0.325 $\pm$ 0.037} & & \cellcolor{teal!20}\textbf{0.506 $\pm$ 0.005}\\
\cmidrule{2-11} \cmidrule{13-13}
& \multirow{5}{*}{0.406} & LaCo & 0.622 & 0.645 & 0.420 & 0.538 & \cellcolor{teal!20}\textbf{0.461} & 0.290 & \cellcolor{teal!20}\textbf{0.334} & 0.232 & & \cellcolor{teal!20}\textbf{0.443}\\
& & LLM-Pruner & 0.594 & 0.658 & 0.431 & 0.534 & 0.395 & \cellcolor{teal!20}\textbf{0.294} & 0.324 & 0.231 & & 0.433\\
& & SliceGPT & 0.378 & 0.581 & 0.404 & 0.579 & 0.443 & 0.270 & 0.296 & 0.230 & & 0.398\\
& & EvoPress & 0.611 & \cellcolor{teal!20}\textbf{0.671} & \cellcolor{teal!20}\textbf{0.457} & 0.541 & 0.443 & 0.274 & 0.280 & 0.251 & & 0.441 \\
& & GeLaCo & \cellcolor{teal!20}\textbf{0.624 $\pm$ 0.002} & 0.606 $\pm$ 0.009 & 0.421 $\pm$ 0.009 & \cellcolor{teal!20}\textbf{0.596 $\pm$ 0.009} & 0.401 $\pm$ 0.003 & \cellcolor{teal!20}\textbf{0.297 $\pm$ 0.003} & 0.307 $\pm$ 0.011 & \cellcolor{teal!20}\textbf{0.303 $\pm$ 0.036} & & \cellcolor{teal!20}\textbf{0.444 $\pm$ 0.005}\\
\cmidrule{1-11} \cmidrule{13-13}
\multirow{18}{*}{\rotatebox[origin=c]{90}{Llama-2-13b}} 
& 0.000 & Dense & 0.805 & 0.805 & 0.794 & 0.720 & 0.774 & 0.492 & 0.452 & 0.521 & & 0.670\\
\cmidrule{2-11} \cmidrule{13-13}
& \multirow{5}{*}{0.150} & LaCo & \cellcolor{teal!20}\textbf{0.686} & 0.768 & 0.746 & 0.706 & 0.703 & 0.462 & \cellcolor{teal!20}\textbf{0.440} & \cellcolor{teal!20}\textbf{0.510} & & 0.627\\
& & LLM-Pruner & 0.676 & \cellcolor{teal!20}\textbf{0.795} & \cellcolor{teal!20}\textbf{0.751} & 0.666 & 0.702 & \cellcolor{teal!20}\textbf{0.474} & 0.434 & 0.272 & & 0.596\\
& & SliceGPT & 0.583 & 0.744 & 0.674 & 0.706 & \cellcolor{teal!20}\textbf{0.718} & 0.451 & 0.430 & 0.360 & & 0.583\\
& & EvoPress & \cellcolor{teal!20}\textbf{0.686} & 0.781 & 0.709 & 0.655 & 0.712 & 0.431 & 0.412 & 0.275 & & 0.583 \\
& & GeLaCo & \cellcolor{teal!20}\textbf{0.732 $\pm$ 0.051} & 0.779 $\pm$ 0.003 & 0.750 $\pm$ 0.000 & \cellcolor{teal!20}\textbf{0.717 $\pm$ 0.003} & 0.707 $\pm$ 0.002 & 0.463 $\pm$ 0.003 & \cellcolor{teal!20}\textbf{0.437 $\pm$ 0.003} & \cellcolor{teal!20}\textbf{0.508 $\pm$ 0.008} & & \cellcolor{teal!20}\textbf{0.637 $\pm$ 0.007}\\
\cmidrule{2-11} \cmidrule{13-13}
& \multirow{5}{*}{0.250} & LaCo & 0.443 & 0.743 & 0.631 & 0.616 & 0.546 & 0.353 & 0.386 & 0.252 & &  0.496\\
& & LLM-Pruner & 0.612 & \cellcolor{teal!20}\textbf{0.764} & 0.668 & 0.609 & 0.614 & 0.405 & \cellcolor{teal!20}\textbf{0.430} & 0.229 & & 0.542\\
& & SliceGPT & 0.430 & 0.691 & 0.581 & 0.684 & 0.621 & \cellcolor{teal!20}\textbf{0.410} & 0.410 & 0.288 & & 0.515\\
& & EvoPress & 0.622 & 0.761 & 0.664 & 0.639 & \cellcolor{teal!20}\textbf{0.671} & 0.398 & 0.396 & 0.298 & & 0.556 \\
& & GeLaCo & \cellcolor{teal!20}\textbf{0.660 $\pm$ 0.020} & 0.742 $\pm$ 0.004 & \cellcolor{teal!20}\textbf{0.693 $\pm$ 0.003} & \cellcolor{teal!20}\textbf{0.696 $\pm$ 0.006} & 0.646 $\pm$ 0.009 & \cellcolor{teal!20}\textbf{0.408 $\pm$ 0.004} & 0.389 $\pm$ 0.006 & \cellcolor{teal!20}\textbf{0.496 $\pm$ 0.007} & & \cellcolor{teal!20}\textbf{0.591 $\pm$ 0.003}\\
\cmidrule{2-11} \cmidrule{13-13}
& \multirow{5}{*}{0.425} & LaCo & \cellcolor{teal!20}\textbf{0.622} & 0.624 & 0.474 & 0.613 & 0.468 & 0.318 & 0.352 & \cellcolor{teal!20}\textbf{0.398} & & 0.484\\
& & LLM-Pruner & 0.396 & 0.638 & 0.394 & 0.523 & 0.306 & 0.255 & 0.310 & 0.235 & & 0.382\\
& & SliceGPT & 0.378 & 0.579 & 0.413 & 0.603 & 0.429 & 0.292 & 0.348 & 0.232 & & 0.409\\
& & EvoPress & 0.617 & \cellcolor{teal!20}\textbf{0.689} & \cellcolor{teal!20}\textbf{0.550} & 0.599 & \cellcolor{teal!20}\textbf{0.513} & \cellcolor{teal!20}\textbf{0.323} & \cellcolor{teal!20}\textbf{0.362} & 0.253 & & \cellcolor{teal!20}\textbf{0.488} \\
& & GeLaCo & \cellcolor{teal!20}\textbf{0.629 $\pm$ 0.010} & 0.627 $\pm$ 0.004 & 0.514 $\pm$ 0.004 & \cellcolor{teal!20}\textbf{0.639 $\pm$ 0.003} & 0.433 $\pm$ 0.008 & \cellcolor{teal!20}\textbf{0.326 $\pm$ 0.006} & 0.312 $\pm$ 0.006 & \cellcolor{teal!20}\textbf{0.401 $\pm$ 0.012} & & \cellcolor{teal!20}\textbf{0.485 $\pm$ 0.003}\\
\cmidrule{1-11} \cmidrule{13-13}
\multirow{18}{*}{\rotatebox[origin=c]{90}{Llama-3.1-8b}} 
& 0.000 & Dense & 0.821 & 0.812 & 0.789 & 0.733 & 0.812 & 0.538 & 0.446 & 0.635 & & 0.698\\
\cmidrule{2-11} \cmidrule{13-13}
& \multirow{5}{*}{0.125} & LaCo & 0.631 & 0.771 & 0.679 & 0.581 & 0.631 & 0.384 & 0.394 & 0.258 & & 0.541\\
& & LLM-Pruner & 0.634 & \cellcolor{teal!20}\textbf{0.793} & 0.726 & 0.703 & 0.718 & 0.445 & 0.410 & 0.411 & & 0.605\\
& & SliceGPT & -- & -- & -- & -- & -- & -- & -- & -- & & -\\
& & EvoPress & \cellcolor{teal!20}\textbf{0.690} & 0.779 & 0.685 & 0.657 & 0.726 & 0.427 & \cellcolor{teal!20}\textbf{0.412} & 0.407 & & 0.598 \\
& & GeLaCo & \cellcolor{teal!20}\textbf{0.696 $\pm$ 0.046} & 0.775 $\pm$ 0.000 & \cellcolor{teal!20}\textbf{0.736 $\pm$ 0.003} & \cellcolor{teal!20}\textbf{0.734 $\pm$ 0.006} & \cellcolor{teal!20}\textbf{0.735 $\pm$ 0.001} & \cellcolor{teal!20}\textbf{0.474 $\pm$ 0.003} & 0.397 $\pm$ 0.001 & \cellcolor{teal!20}\textbf{0.604 $\pm$ 0.004} & & \cellcolor{teal!20}\textbf{0.644 $\pm$ 0.006}\\
\cmidrule{2-11} \cmidrule{13-13}
& \multirow{5}{*}{0.187} & LaCo & 0.498 & 0.679 & 0.530 & 0.595 & 0.520 & 0.343 & 0.344 & 0.253 & & 0.470\\
& & LLM-Pruner & 0.568 & \cellcolor{teal!20}\textbf{0.772} & \cellcolor{teal!20}\textbf{0.687} & 0.684 & 0.657 & 0.402 & \cellcolor{teal!20}\textbf{0.392} & 0.277 & & 0.555\\
& & SliceGPT & -- & -- & -- & -- & -- & -- & -- & -- & & -\\
& & EvoPress & 0.551 & 0.752 & 0.634 & 0.587 & 0.647 & 0.358 & 0.370 & 0.276 & & 0.522 \\
& & GeLaCo & \cellcolor{teal!20}\textbf{0.760 $\pm$ 0.019} & 0.734 $\pm$ 0.005 & 0.678 $\pm$ 0.003 & \cellcolor{teal!20}\textbf{0.709 $\pm$ 0.004} & \cellcolor{teal!20}\textbf{0.679 $\pm$ 0.011} & \cellcolor{teal!20}\textbf{0.435 $\pm$ 0.007} & 0.360 $\pm$ 0.012 & \cellcolor{teal!20}\textbf{0.597 $\pm$ 0.005} & & \cellcolor{teal!20}\textbf{0.619 $\pm$ 0.003}\\
\cmidrule{2-11} \cmidrule{13-13}
& \multirow{5}{*}{0.312} & LaCo & 0.395 & 0.610 & 0.370 & 0.566 & 0.404 & 0.282 & 0.302 & 0.243 & & 0.397\\
& & LLM-Pruner & \cellcolor{teal!20}\textbf{0.499} & \cellcolor{teal!20}\textbf{0.706} & 0.454 & 0.564 & 0.487 & 0.286 & 0.302 & 0.229 & & 0.441\\
& & SliceGPT & -- & -- & -- & -- & -- & -- & -- & -- & & -\\
& & EvoPress & 0.435 & 0.700 & \cellcolor{teal!20}\textbf{0.533} & 0.559 & \cellcolor{teal!20}\textbf{0.549} & \cellcolor{teal!20}\textbf{0.309} & \cellcolor{teal!20}\textbf{0.340} & 0.241 & & \cellcolor{teal!20}\textbf{0.458} \\
& & GeLaCo & \cellcolor{teal!20}\textbf{0.495 $\pm$ 0.050} & 0.666 $\pm$ 0.005 & 0.507 $\pm$ 0.018 & \cellcolor{teal!20}\textbf{0.612 $\pm$ 0.027} & 0.485 $\pm$ 0.024 & \cellcolor{teal!20}\textbf{0.315 $\pm$ 0.013} & 0.306 $\pm$ 0.002 & \cellcolor{teal!20}\textbf{0.303 $\pm$ 0.023} & & \cellcolor{teal!20}\textbf{0.461 $\pm$ 0.009}\\
\bottomrule
\end{tabular}}

\caption{Benchmark performance across models and compression methods at specific compression ratios. Highlighted bold values indicate the best performing method for each task within each model and compression ratio.}
\label{tab:ppl_performance}
\end{table*}

We first evaluated GeLaCo under the single-objective formulation, aiming to find the best possible configuration given a specific compression target. The results are shown in Table ~\ref{tab:ppl_performance}, comparing GeLaCo with the selected baselines across the three selected foundational models: Llama-2 7B, Llama-2 13B, and Llama-3.1 8B. 

Since LaCo does not directly specify target compression ratios, but rather determines the compression outcome empirically, we establish the compression ratios achieved by LaCo runs as target objectives for all competing methods.\footnote{Due to differences in our experimental setup, mainly calibration data sampling, the compression ratios we achieved with LaCo differ slightly from those reported in their work.} For Llama-3.1 8B, we evaluated LaCo's portability using the pre-established LaCo hyper-parameters for the closest Llama-2 model in size, Llama-2 7B, as no equivalent was available for Llama-3 models.

The first notable result is that, considering the average across benchmarks, GeLaCo is either the top-performing approach, in six out of nine configurations, or tied with a competing method, different in all three such cases. Our parameterized approach to layer collapse, along with the selected fitness function, thus seem to provide a robust basis for compression solution search across scenarios, at least for the Llama model family.

Another relevant observation is that the best-performing methods can vary depending on the benchmark at hand, showcasing the difficulties in reaching compression variants that can systematically reach optimal quality preservation across all different tasks. Thus, although GeLaCo provides a more balanced behaviour, as shown by average results, it underperforms slightly but systematically on tasks such as PIQA or Hellaswag. For other tasks, such as BoolQ, Winogrande or MMLU, it achieves the best performance in virtually all cases. 

Varying performance degradation depending on the task might be due to the reliance on specific parts of the network for the resolution of a given task, dependent on the distribution of the relevant knowledge within deep networks. Despite the observed variance in specific configurations, model type, size and compression ratio, when considering all tasks, GeLaCo achieved the highest scores (ties included) in 52.8\% of the cases, followed by LLM-Pruner at 27.8\%, EvoPress at 22.2\%, LaCo at 15.2\% and SliceGPT at 4.2\%. 

Of note are the comparative results overall between GeLaCo and EvoPress, two structured pruning evolutionary approaches. Although the latter can provide higher scoring compression solutions in some of the configurations, particularly at the highest compression ratios for Llama3, where it achieved the best results in four of the eight benchmarks, it appears to be more affected than the former as model type, size and compression ratios vary. Including parametrized layer contributions, as in the GeLaCo approach, thus seems to provide clear benefits in terms of robustness of evolutionary compression solutions. 

Finally, it is also worth noting that GeLaCo demonstrates balanced performance across the evaluation suite while maintaining stable optimization behavior, as indicated by the minor standard deviations across the three runs performed in each compression scenario. 

\subsection{Bi-objective Optimization}

\begin{figure*}[t!]
    \centering
    \resizebox{\textwidth}{!}{
    \begin{tabular}{cc}
        \includegraphics[width=0.5\textwidth]{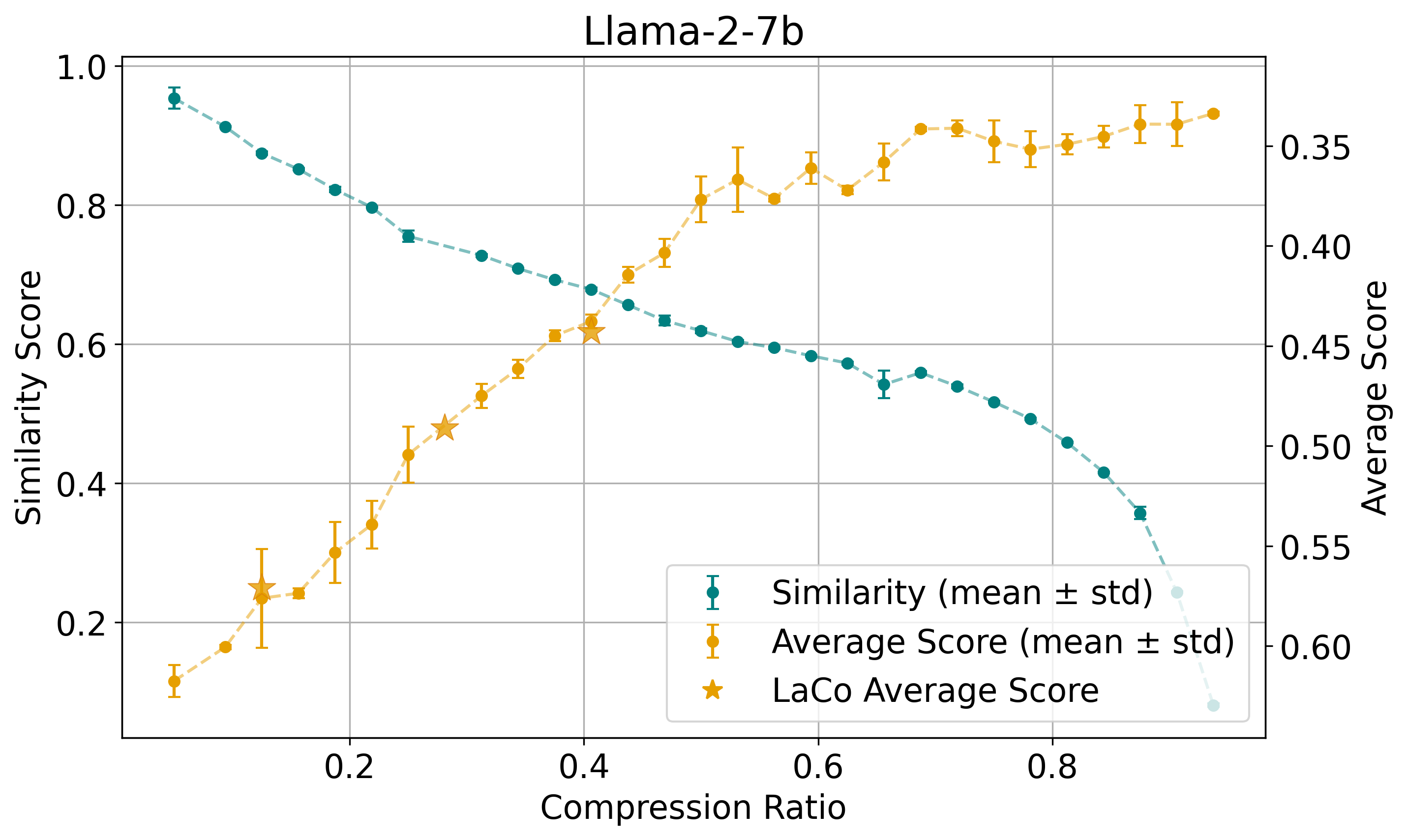} & 
        \includegraphics[width=0.5\textwidth]{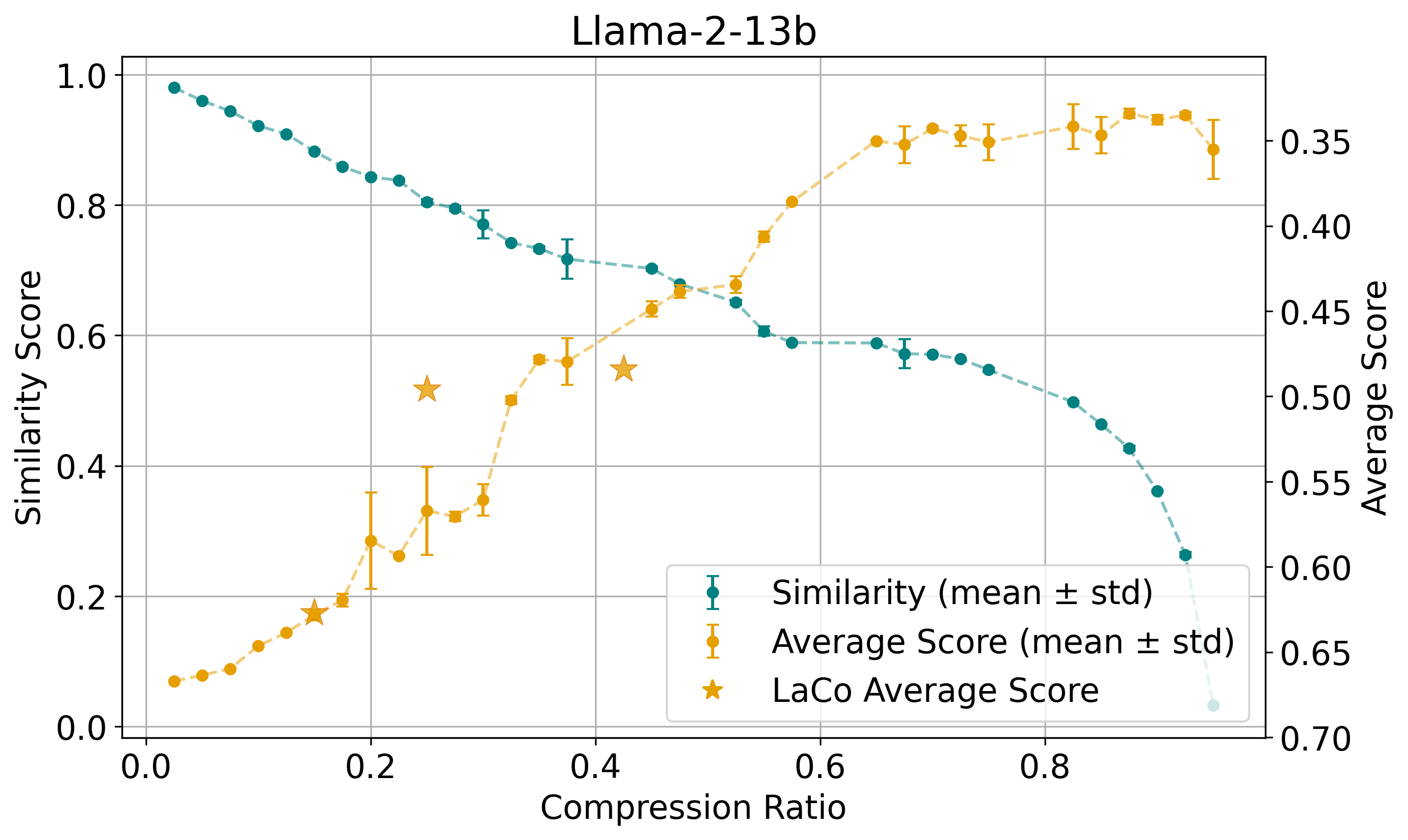} \\[0em]
        \includegraphics[width=0.5\textwidth]{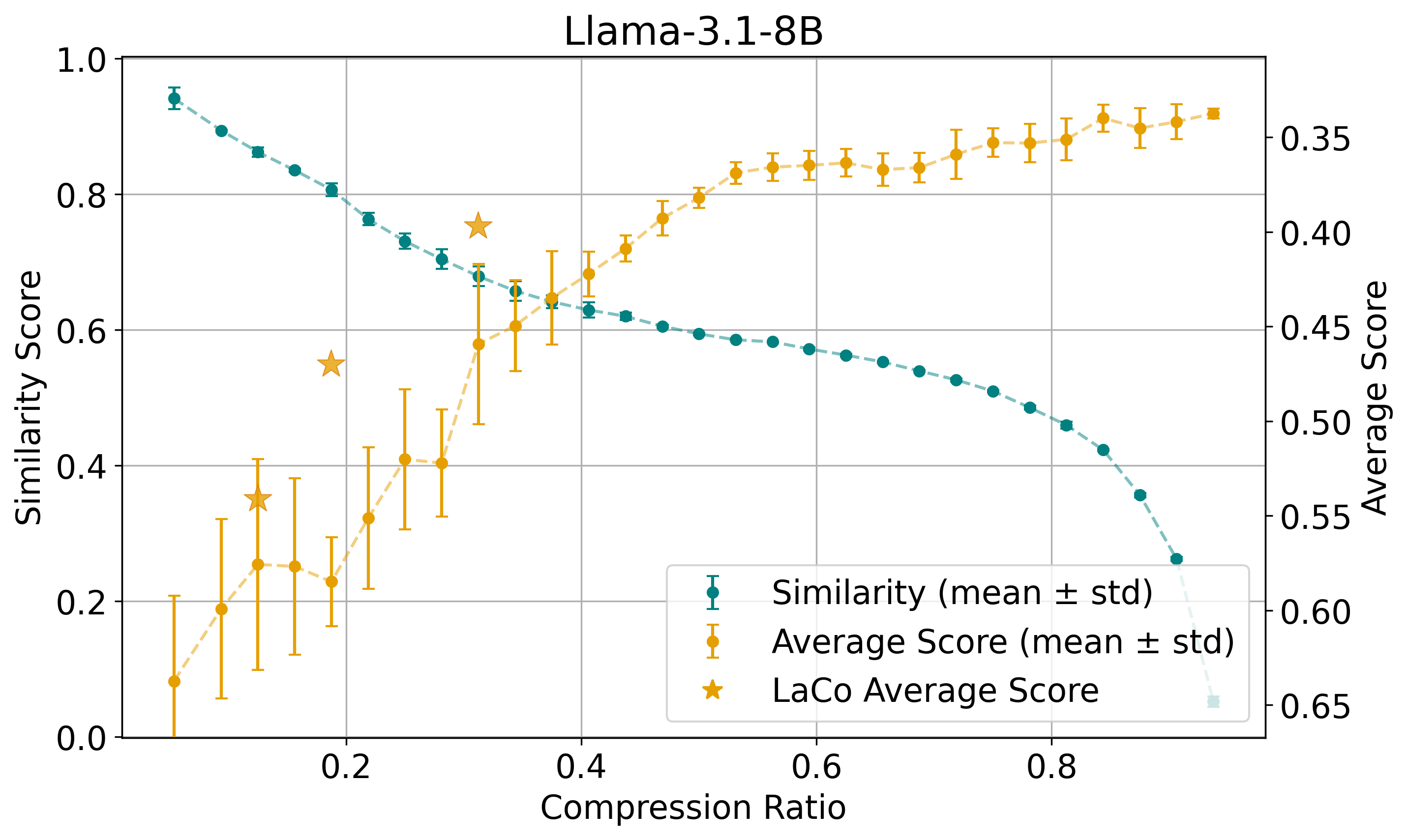} & 
        \includegraphics[width=0.5\textwidth]{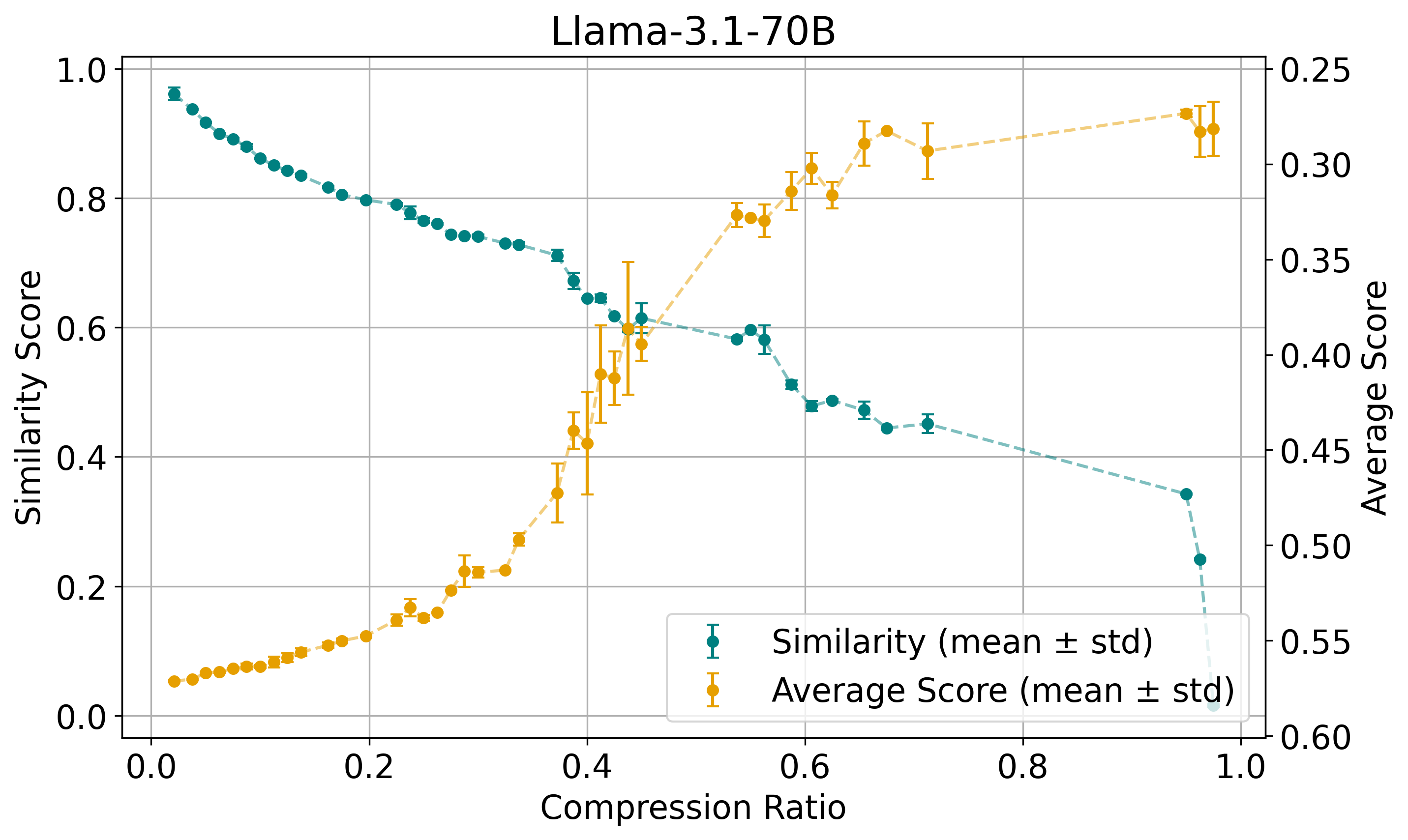}
    \end{tabular}}
    \caption{Mean Pareto front approximation for dual objective optimization across different Llama models showing fitness value and average performance over all benchmarks. Top row: Llama-2 7B (left) and Llama-2 13B (right). Bottom row: Llama-3.1 8B (left) and Llama-3.1 70B (right).}
    \label{fig:pareto_comparison}
\end{figure*}

For these experiments, we selected compression ratio and model similarity as dual objectives. Figure~\ref{fig:pareto_comparison} presents a comparison of the Pareto front approximations estimated by GeLaCo for the selected Llama models, when optimizing for both compression and quality objectives. We include in the Figure the compression ratios, similarity scores and average scores over all benchmarks.\footnote{Note that the similarity and quality scales are inverted in the graph: bottom (worst) to top (best) for the former, and the reverse for the latter.} Whiskers represent the variability of the similarity and score among the points in the estimated Pareto fronts computed across three independent runs for every experiment. We also indicate LaCo solutions at specific compression rates, to situate them within the Pareto front. 

For these experiments, we extended our analysis to the larger Llama-3.1 70B model, as compressing larger models, while preserving the quality achieved at larger scales, poses specific challenges. For this model, we do not include LaCo solutions, as there are no pre-established nor portable configurations for a model with a larger number of layers than the Llama-2 models.

For all models, we observe similar patterns, with decreasing similarity and quality scores as the compression ratio increases. This is not unexpected, as compression affects model capacity. However, the multi-objective GeLaCo approach enables a precise characterization of the trade-offs between compression and quality. 

Overall, GeLaCo solutions dominate, in the Pareto sense, those elicited by LaCo. For Llama-2 7B, results are relatively similar between the two approaches in this respect; for Llama-2 13B, one LaCo solution achieved relatively higher quality, one was virtually tied, and the third comparable data point showed a significantly higher-scoring GeLaCo solution; for LLama-3 8B, all GeLaco solutions achieved substantial gains over comparable LaCo solutions. Note that, for Llama-2 13B, the single-objective results in Table~\ref{tab:ppl_performance} indicate that GeLaCo is able to find a better solution than LaCo at 0.425 compression. The single case where a LaCo solution dominates the Pareto front can be due to the limited search budget in the bi-objective optimization, but in no case to a limitation of the GeLaCo approach.
 
The similarity metric (shown in teal) demonstrates a characteristic progression across compression ratios for all models: scores decrease gradually until approximately 0.6 compression ratio, and then exhibit a relatively stable plateau. This plateau extends until a compression level around 0.9 before dropping toward zero at the most extreme compression levels. The Llama-3.1 70B model incurs even more erratic similarity drops beyond the 0.6 compression mark, indicating specific behavioural characteristics of deeper models when affected by higher compression.

\begin{table*}[t]
\footnotesize
\centering
\resizebox{\textwidth}{!}{%
\begin{tabular}{lcccccccccc}
\toprule
\textbf{Method} & \textbf{boolq} & \textbf{piqa} & \textbf{hellaswag} & \textbf{winogrande} & \textbf{arc\_easy} & \textbf{arc\_challenge} & \textbf{openbookqa} & \textbf{mmlu} & & \textbf{average} \\
\cmidrule{1-9} \cmidrule{11-11}
Dense & 0.821 & 0.812 & 0.789 & 0.733 & 0.812 & 0.538 & 0.446 & 0.635 & & 0.698 \\
\cmidrule{1-9} \cmidrule{11-11}
GeLaCo & 0.623 $\pm$ 0.000 & 0.556 $\pm$ 0.001 & 0.314 $\pm$ 0.004 & 0.553 $\pm$ 0.002 & 0.323 $\pm$ 0.004 & 0.276 $\pm$ 0.000 & 0.302 $\pm$ 0.006 & 0.231 $\pm$ 0.001 & & 0.397 $\pm$ 0.001 \\
GeLaCo (2.5B) & \cellcolor{teal!20}\textbf{0.715} $\pm$ 0.011 & 0.730 $\pm$ 0.001 & 0.619 $\pm$ 0.003 & \cellcolor{teal!20}\textbf{0.650} $\pm$ 0.006 & 0.663 $\pm$ 0.004 & 0.402 $\pm$ 0.013 & 0.383 $\pm$ 0.009 & 0.307 $\pm$ 0.013 & & 0.559 $\pm$ 0.003 \\
GeLaCo (5B) & 0.675 $\pm$ 0.029 & 0.746 $\pm$ 0.003 & 0.643 $\pm$ 0.003 & \cellcolor{teal!20}\textbf{0.648} $\pm$ 0.002 & \cellcolor{teal!20}\textbf{0.671} $\pm$ 0.002 & 0.423 $\pm$ 0.006 & \cellcolor{teal!20}\textbf{0.390} $\pm$ 0.002 & \cellcolor{teal!20}\textbf{0.322} $\pm$ 0.010 & & 0.565 $\pm$ 0.004 \\
GeLaCo (7.5B) & \cellcolor{teal!20}\textbf{0.706} $\pm$ 0.015 & 0.745 $\pm$ 0.003 & 0.658 $\pm$ 0.002 & \cellcolor{teal!20}\textbf{0.655} $\pm$ 0.003 & \cellcolor{teal!20}\textbf{0.673} $\pm$ 0.006 & 0.425 $\pm$ 0.006 & \cellcolor{teal!20}\textbf{0.394} $\pm$ 0.000 & 0.363 $\pm$ 0.014 & & \cellcolor{teal!20}\textbf{0.577} $\pm$ 0.003 \\
GeLaCo (10B) & \cellcolor{teal!20}\textbf{0.708} $\pm$ 0.010 & \cellcolor{teal!20}\textbf{0.752} $\pm$ 0.002 & 0.667 $\pm$ 0.001 & \cellcolor{teal!20}\textbf{0.654} $\pm$ 0.005 & \cellcolor{teal!20}\textbf{0.688} $\pm$ 0.002 & \cellcolor{teal!20}\textbf{0.437} $\pm$ 0.005 & \cellcolor{teal!20}\textbf{0.404} $\pm$ 0.000 & \cellcolor{teal!20}\textbf{0.376} $\pm$ 0.012 & & \cellcolor{teal!20}\textbf{0.586} $\pm$ 0.002 \\
\cmidrule{1-9} \cmidrule{11-11}
DarwinLM (10B) & 0.710 & 0.748 & 0.669 & 0.613 & 0.671 & 0.434 & 0.396 & 0.375 & & 0.577 \\
\bottomrule
\end{tabular}%
}
\caption{Post-training performance of the Llama-3.1 8B model compressed at 50\%, with different volumes of post-training data. Results that match or outperform the DarwinLM 10B baseline are shown in bold.}
\label{tab:post-training_performance}
\end{table*}

Average benchmark performance (in yellow) follows a similar decreasing trend, with relatively steady losses as compression increases, up to the 0.6 mark, where quality results start tending towards near-random levels. Models of comparable size, namely Llama-2 7B and Llama-3.1, featured a relatively similar evolution of their scores as compression increased, whereas the larger Llama-2 13B started exhibiting non-monotonic quality results after the 0.5 compression mark, a trend which is significantly accentuated with the larger Llama-3.1 70B model. This tendency can be likely attributed to the models' larger sizes, where the same compression ratio necessitates removing a greater absolute number of layers, resulting in more substantial alterations to the models' internal parameter distribution and activation pathways.

Although our similarity metric can serve as a reasonable proxy for knowledge preservation considering the Pareto results, it will be worth considering alternative approaches to model similarity in future work, to establish a closer pairing between fitness scoring and model quality.

\subsection{Post-training}
\label{sec:post-training}

As previously indicated, compressed models can strongly benefit from a post-training phase, in the form of continued next-token prediction training over raw text and/or instruction datasets. We provide results along these lines, first over base models, then over instruction-tuned variants.

\paragraph{Foundational Models.}  In Table~\ref{tab:post-training_performance} we present the post-training results of GeLaCo over base models, including as a reference the results for DarwinLM \citet{tang2025darwinlm} with a post-training dataset of 10B tokens. The DarwinLM model contains 4.6B parameters, while our GeLaCo compressed model contains 4.5B parameters, and are thus comparable in terms of model size. For GeLaCo, we fine-tuned the compressed models using the AdamW optimizer with learning rate equal to 2e-5, and an effective batch size of 128. We gradually increased the amount of data from the Fineweb-Edu dataset up to 10B tokens, to measure progressive improvements via post-training regularization.

The results in Table~\ref{tab:post-training_performance} show gradual improvements across all benchmarks as data volumes increase, with occasional minor losses on specific benchmarks for new model iterations, e.g. GeLaCo (5B) on Boolq. Remarkably, GeLaCo (7.5B) against DarwinLM (10B) achieve similar averages, although the differences are relatively small and additional baselines for the latter would be needed for a fair comparison.

\begin{table*}[t]
\centering
\resizebox{2\columnwidth}{!}{\begin{tabular}{ccccccccccccc}
\toprule
\textbf{Model} & \textbf{Ratio} & \textbf{Method} & \textbf{boolq} & \textbf{piqa} & \textbf{hellaswag} & \textbf{winogrande} & \textbf{arc\_easy} & \textbf{arc\_challenge} & \textbf{openbookqa} & \textbf{mmlu} & & \textbf{average} \\
\cmidrule{1-11} \cmidrule{13-13}
\multirow{8}{*}{\rotatebox[origin=c]{90}{Qwen3-8b}} 
& 0.000 & Dense & 0.866 & 0.779 & 0.748 & 0.677 & 0.808 & 0.567 & 0.414 & 0.730 & & 0.699 \\
\cmidrule{2-11} \cmidrule{13-13}
& \multirow{2}{*}{0.222} & EvoPress & \cellcolor{teal!20}\textbf{0.705} & \cellcolor{teal!20}\textbf{0.737} & \cellcolor{teal!20}\textbf{0.592} & \cellcolor{teal!20}\textbf{0.593} & \cellcolor{teal!20}\textbf{0.721} & \cellcolor{teal!20}\textbf{0.452} & \cellcolor{teal!20}\textbf{0.398} & \cellcolor{teal!20}\textbf{0.461} & & \cellcolor{teal!20}\textbf{0.582} \\
& & GeLaCo & 0.687 $\pm$ 0.014 & 0.700 $\pm$ 0.013 & 0.533 $\pm$ 0.015 & 0.583 $\pm$ 0.009 & 0.602 $\pm$ 0.012 & 0.380 $\pm$ 0.012 & 0.354 $\pm$ 0.010 & \cellcolor{teal!20}\textbf{0.451} $\pm$ 0.035 & & 0.536 $\pm$ 0.006 \\
\cmidrule{2-11} \cmidrule{13-13}
& \multirow{2}{*}{0.417} & EvoPress & 0.407 & \cellcolor{teal!20}\textbf{0.635} & \cellcolor{teal!20}\textbf{0.385} & 0.511 & \cellcolor{teal!20}\textbf{0.480} & \cellcolor{teal!20}\textbf{0.288} & \cellcolor{teal!20}\textbf{0.318} & \cellcolor{teal!20}\textbf{0.252} & & 0.410 \\
& & GeLaCo & \cellcolor{teal!20}\textbf{0.611} $\pm$ 0.010 & 0.597 $\pm$ 0.006 & 0.356 $\pm$ 0.007 & \cellcolor{teal!20}\textbf{0.536} $\pm$ 0.007 & 0.449 $\pm$ 0.018 & 0.275 $\pm$ 0.002 & 0.299 $\pm$ 0.003 & \cellcolor{teal!20}\textbf{0.247} $\pm$ 0.008 & & \cellcolor{teal!20}\textbf{0.421} $\pm$ 0.003 \\
\cmidrule{2-11} \cmidrule{13-13}
& \multirow{2}{*}{0.583} & EvoPress & 0.404 & \cellcolor{teal!20}\textbf{0.561} & \cellcolor{teal!20}\textbf{0.302} & 0.511 & \cellcolor{teal!20}\textbf{0.347} & 0.248 & \cellcolor{teal!20}\textbf{0.288} & 0.230 & & 0.361 \\
& & GeLaCo & \cellcolor{teal!20}\textbf{0.614} $\pm$ 0.007 & 0.530 $\pm$ 0.002 & 0.287 $\pm$ 0.001 & \cellcolor{teal!20}\textbf{0.517} $\pm$ 0.005 & 0.259 $\pm$ 0.001 & \cellcolor{teal!20}\textbf{0.271} $\pm$ 0.001 & 0.269 $\pm$ 0.005 & \cellcolor{teal!20}\textbf{0.256} $\pm$ 0.004 & & \cellcolor{teal!20}\textbf{0.375} $\pm$ 0.001 \\
\bottomrule
\end{tabular}}
\caption{Benchmark performance for Qwen3 8B across compression methods at specific compression ratios. Highlighted bold values indicate the best performing method for each task within each compression ratio.}
\label{tab:cross_architecture}
\end{table*}

\begin{figure}[h]
    \centering
    \includegraphics[width=\linewidth]{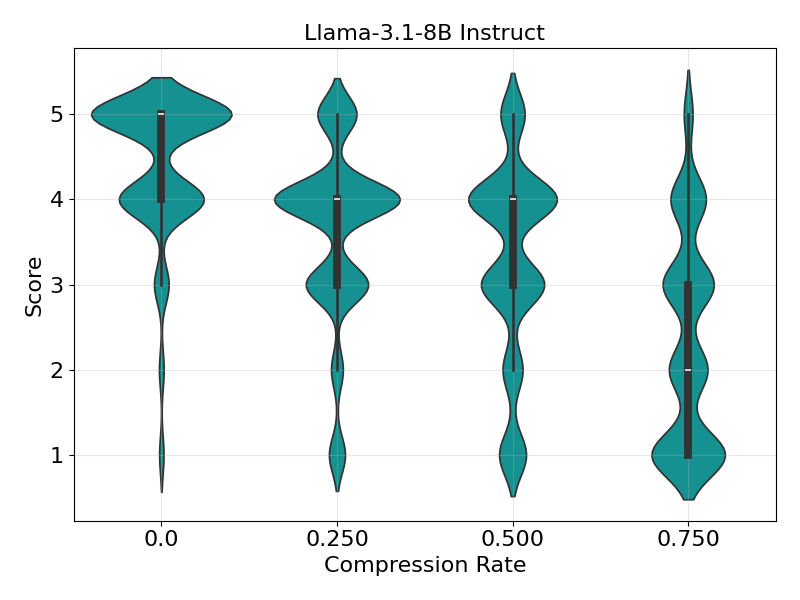}
    \caption{Violin plots showing the distribution of instruction-following performance scores for Llama-3.1 8B Instruct across different compression rates. The central dot in each violin represents the median, whereas the rectangle shows the interquartile range.}
    \label{fig:instruction-results}
\end{figure}

\paragraph{Instruction Models.}  Figure ~\ref{fig:instruction-results} illustrates the distribution of instruction-following performance scores for Llama-3.1 8B Instruct across different compression rates after post-training recovery using the LaMini Instruction dataset, evaluated on the Just-Eval dataset using Prometheus 2 as a judge. We fine-tune compressed models for 2 epochs using the AdamW optimizer with learning rate 2e-5 and an effective batch size of 128.

At low to moderate compression rates, GeLaCo maintains strong instruction-following performance after recovery fine-tuning. At 0.25 compression, the model achieves a median score of 4 with score distributions concentrated in the 3-4 range. At 0.5 compression, performance remains robust with a median score of 4 and similar distribution patterns, demonstrating that meaningful size reductions can be accomplished and effectively recovered without significantly compromising specialized instruction-following capabilities.

At more aggressive compression rates (0.75), we observe a clear drop in instruction-following quality, with the distribution shifting towards lower scores. Notably, the distribution becomes significantly more variable, spanning the full range from 1-4 scores, revealing increased performance inconsistency under high compression. Nevertheless, the compressed models retains some instruction-following capabilities, as evidenced by the distribution's upper quartile still reaching a score of 3 and a median of 2. While performance becomes more constrained at this compression level, these results establish the practical boundaries of layer collapse compression for instruction-tuned models and demonstrate that even under extreme compression, some instruction-following capabilities can be preserved.

\subsection{Cross-Architecture Evaluation}

To evaluate the portability of our approach to another architecture, we compressed Qwen3 8B under three representative settings covering increasingly higher compression ratios. Table~\ref{tab:cross_architecture} reports the corresponding single-objective results. For the Llama family, GeLaCo outperformed EvoPress solutions in virtually all cases. However, on Qwen3 8B the results are more mixed. EvoPress achieves the best average at the lowest compression ratio of 0.222, and the highest scores across all downstream tasks except MMLU, where both approaches are tied. At higher compression rates, GeLaCo achieves the best average over all tasks, although EvoPress remains competitive with higher scores on several downstream tasks. GeLaCo tends to underperform slightly but systematically on tasks such as PIQA and HellaSwag, while performing better on BoolQ, Winogrande, and MMLU, reinforcing the tendencies observed in Section~\ref{sec:single-objective}. 

Overall, GeLaCo shows clear portability to a different architecture while suggesting that its robustness is less uniformly dominant on Qwen3 8B than on the Llama family. Further study will be needed to determine whether this behavior reflects properties of the Qwen architecture itself or a broader interaction between architecture and compression strategy.

\section{Conclusions}

We presented GeLaCo, an efficient evolutionary approach to language model compression featuring a novel layer collapse framework based on parametrized weight merging, and a novel fitness function which jointly computes similarity over residual updates and KL divergence. 

In terms of single-objective optimization, with fixed target compression ratios, GeLaCo outperformed state-of-the-art alternatives overall, namely LaCo, LLM-Pruner, SliceGPT and EvoPress. It achieved the best average scores over the Llama-2 and Llama-3 models, also leading all approaches on the majority of benchmarks. Evaluations on the Qwen3 8B model showed GeLaCo remained competitive beyond the Llama family.

Multi-objective optimization led to establishing the first Pareto front estimation in terms of language model compression and quality, where GeLaCo solutions were Pareto dominant in most cases when compared to the original LaCo solutions. These results  showcased the challenges presented by models of different sizes, with sharp quality declines beyond 50\% compression rates for most models, specifically the larger variants such as Llama-3.1 70B.

Post-training GeLaCo solutions resulted in linear improvements overall as post-training data volumes increased, surpassing the state-of-the-art baseline overall. We also provided the first generative results for instruction-tuned compressed models, showing that moderate to medium compression could preserve a significant proportion of answer quality for post-trained compression models. 

Finally, since GeLaCo preserves the standard Transformer block structure, its compressed checkpoints remain compatible with optimized inference engines, making it immediately deployable  without requiring model-specific kernels or custom inference code.


Although our fitness similarity function outperformed alternatives based on KL divergence or perplexity, exploring novel similarity functions will be an important future research path, to establish a closer pairing between fitness scoring and model quality. Additionally, it will be worth considering alternative EA approaches that prioritize population diversity to enhance evolutionary search and avoid population stagnation.

\section*{Acknowledgments}
This work was partially supported by the Department of Economic Development and Competitiveness of the Basque Government (SPRI Group) through funding for project IKUN (KK2024/00064). J. Del Ser also acknowledges funding support from the Basque Government through the consolidated research group MATHMODE (IT1456-22).


\bibliographystyle{acl_natbib}
\bibliography{anthology,custom}

@article{tang2025darwinlm,
  title={DarwinLM: Evolutionary Structured Pruning of Large Language Models},
  author={Tang, Shengkun and Sieberling, Oliver and Kurtic, Eldar and Shen, Zhiqiang and Alistarh, Dan},
  journal={arXiv preprint arXiv:2502.07780},
  year={2025}
}

@article{sieberling2024evopress,
  title={{EvoPress}: Towards optimal dynamic model compression via evolutionary search},
  author={Sieberling, Oliver and Kuznedelev, Denis and Kurtic, Eldar and Alistarh, Dan},
  journal={arXiv preprint arXiv:2410.14649},
  year={2024}
}

@article{huang2025towards,
  title={Towards Efficient Automatic Self-Pruning of Large Language Models},
  author={Huang, Weizhong and Zhang, Yuxin and Zheng, Xiawu and Chao, Fei and Ji, Rongrong},
  journal={arXiv preprint arXiv:2502.14413},
  year={2025}
}

@inproceedings{xiao2023smoothquant,
  title={Smoothquant: Accurate and efficient post-training quantization for large language models},
  author={Xiao, Guangxuan and Lin, Ji and Seznec, Mickael and Wu, Hao and Demouth, Julien and Han, Song},
  booktitle={International Conference on Machine Learning},
  pages={38087--38099},
  year={2023},
  organization={PMLR}
}

@article{lin2024awq,
  title={{AWQ}: Activation-aware weight quantization for on-device {LLM} compression and acceleration},
  author={Lin, Ji and Tang, Jiaming and Tang, Haotian and Yang, Shang and Chen, Wei-Ming and Wang, Wei-Chen and Xiao, Guangxuan and Dang, Xingyu and Gan, Chuang and Han, Song},
  journal={Proceedings of Machine Learning and Systems},
  volume={6},
  pages={87--100},
  year={2024}
}

@article{wang2023bitnet,
  title={{BitNet}: Scaling 1-bit transformers for large language models},
  author={Wang, Hongyu and Ma, Shuming and Dong, Li and Huang, Shaohan and Wang, Huaijie and Ma, Lingxiao and Yang, Fan and Wang, Ruiping and Wu, Yi and Wei, Furu},
  journal={arXiv preprint arXiv:2310.11453},
  year={2023}
}

@inproceedings{huang2024billm,
  title={{BiLLM}: pushing the limit of post-training quantization for {LLMs}},
  author={Huang, Wei and Liu, Yangdong and Qin, Haotong and Li, Ying and Zhang, Shiming and Liu, Xianglong and Magno, Michele and Qi, Xiaojuan},
  booktitle={International Conference on Machine Learning},
  pages={20023--20042},
  year={2024}
}

@article{xu2024onebit,
  title={{OneBit}: Towards extremely low-bit large language models},
  author={Xu, Yuzhuang and Han, Xu and Yang, Zonghan and Wang, Shuo and Zhu, Qingfu and Liu, Zhiyuan and Liu, Weidong and Che, Wanxiang},
  journal={arXiv preprint arXiv:2402.11295},
  year={2024}
}

@article{ma2024era,
  title={The Era of 1-bit {LLMs}: All Large Language Models are in 1.58 Bits},
  author={Ma, Shuming and Wang, Hongyu and Ma, Lingxiao and Wang, Lei and Wang, Wenhui and Huang, Shaohan and Dong, Li and Wang, Ruiping and Xue, Jilong and Wei, Furu},
  journal={arXiv preprint arXiv:2402.17764},
  year={2024}
}

@article{liu2024spinquant,
  title={{SpinQuant: LLM quantization with learned rotations}},
  author={Liu, Zechun and Zhao, Changsheng and Fedorov, Igor and Soran, Bilge and Choudhary, Dhruv and Krishnamoorthi, Raghuraman and Chandra, Vikas and Tian, Yuandong and Blankevoort, Tijmen},
  journal={arXiv preprint arXiv:2405.16406},
  year={2024}
}

@inproceedings{liu2024llm,
  title={{LLM-QAT}: Data-Free Quantization Aware Training for Large Language Models},
  author={Liu, Zechun and Oguz, Barlas and Zhao, Changsheng and Chang, Ernie and Stock, Pierre and Mehdad, Yashar and Shi, Yangyang and Krishnamoorthi, Raghuraman and Chandra, Vikas},
  booktitle={Findings of the Association for Computational Linguistics ACL 2024},
  pages={467--484},
  year={2024}
}

@article{men2024shortgpt,
  title={{ShortGPT}: Layers in large language models are more redundant than you expect},
  author={Men, Xin and Xu, Mingyu and Zhang, Qingyu and Wang, Bingning and Lin, Hongyu and Lu, Yaojie and Han, Xianpei and Chen, Weipeng},
  journal={arXiv preprint arXiv:2403.03853},
  year={2024}
}

@article{vaswani2017attention,
  title={Attention is all you need},
  author={Vaswani, Ashish and Shazeer, Noam and Parmar, Niki and Uszkoreit, Jakob and Jones, Llion and Gomez, Aidan N and Kaiser, {\L}ukasz and Polosukhin, Illia},
  journal={Advances in neural information processing systems},
  volume={30},
  year={2017}
}

@inproceedings{frantar2023sparsegpt,
  title={{SparseGPT}: Massive language models can be accurately pruned in one-shot},
  author={Frantar, Elias and Alistarh, Dan},
  booktitle={International Conference on Machine Learning},
  pages={10323--10337},
  year={2023},
  organization={PMLR}
}

@article{xia2023sheared,
  title={Sheared {Llama}: Accelerating language model pre-training via structured pruning},
  author={Xia, Mengzhou and Gao, Tianyu and Zeng, Zhiyuan and Chen, Danqi},
  journal={arXiv preprint arXiv:2310.06694},
  year={2023}
}

@article{touvron2023llama,
  title={Llama 2: Open foundation and fine-tuned chat models},
  author={Touvron, Hugo and others},
  journal={arXiv preprint arXiv:2307.09288},
  year={2023}
}

@inproceedings{cao2024head,
  title={Head-wise Shareable Attention for Large Language Models},
  author={Cao, Zouying and Yang, Yifei and Zhao, Hai},
  booktitle={Findings of the Association for Computational Linguistics: EMNLP 2024},
  pages={2555--2571},
  year={2024}
}

@article{dettmers2023spqr,
  title={{SpQR}: A sparse-quantized representation for near-lossless {LLM} weight compression},
  author={Dettmers, Tim and Svirschevski, Ruslan and Egiazarian, Vage and Kuznedelev, Denis and Frantar, Elias and Ashkboos, Saleh and Borzunov, Alexander and Hoefler, Torsten and Alistarh, Dan},
  journal={arXiv preprint arXiv:2306.03078},
  year={2023}
}

@article{ma2023llm,
  title={{LLM-pruner}: On the structural pruning of large language models},
  author={Ma, Xinyin and Fang, Gongfan and Wang, Xinchao},
  journal={Advances in Neural Information Processing Systems},
  volume={36},
  pages={21702--21720},
  year={2023}
}

@article{ashkboos2024slicegpt,
  title={{SliceGPT}: Compress large language models by deleting rows and columns},
  author={Ashkboos, Saleh and Croci, Maximilian L and Nascimento, Marcelo Gennari do and Hoefler, Torsten and Hensman, James},
  journal={arXiv preprint arXiv:2401.15024},
  year={2024}
}

@article{benitez2019jmetalpy,
  title={{jMetalPy}: A {Python} framework for multi-objective optimization with metaheuristics},
  author={Ben{\'\i}tez-Hidalgo, Antonio and Nebro, Antonio J and Garc{\'\i}a-Nieto, Jos{\'e} and Oregi, Izaskun and Del Ser, Javier},
  journal={Swarm and Evolutionary Computation},
  volume={51},
  pages={100598},
  year={2019},
  publisher={Elsevier}
}

@article{grattafiori2024llama,
  title={The {Llama 3} herd of models},
  author={Grattafiori, Aaron and others},
  journal={arXiv preprint arXiv:2407.21783},
  year={2024}
}

@article{wu2025evop,
  title={{EvoP}: Robust {LLM} Inference via Evolutionary Pruning},
  author={Wu, Shangyu and Du, Hongchao and Xiong, Ying and Chen, Shuai and Kuo, Tei-wei and Guan, Nan and Xue, Chun Jason},
  journal={arXiv preprint arXiv:2502.14910},
  year={2025}
}

@misc{lozhkov2024fineweb-edu,
    author       = { Lozhkov, Anton and Ben Allal, Loubna and von Werra, Leandro and Wolf, Thomas },  
    title        = { FineWeb-Edu: the Finest Collection of Educational Content }, 
    year         = 2024,  
    url          = { https://huggingface.co/datasets/HuggingFaceFW/fineweb-edu },  
    doi          = { 10.57967/hf/2497 },
    publisher    = { Hugging Face }
}

@article{wu2023lamini-lm,
  author       = {Minghao Wu and
                  Abdul Waheed and
                  Chiyu Zhang and
                  Muhammad Abdul-Mageed and
                  Alham Fikri Aji
                  },
  title        = {LaMini-LM: A Diverse Herd of Distilled Models from Large-Scale Instructions},
  journal      = {CoRR},
  volume       = {abs/2304.14402},
  year         = {2023},
  url          = {https://arxiv.org/abs/2304.14402},
  eprinttype   = {arXiv},
  eprint       = {2304.14402}
}

@article{Lin2023ReAlign,
  title={The unlocking spell on base {LLMs}: Rethinking alignment via in-context learning},
  author={Lin, Bill Yuchen and Ravichander, Abhilasha and Lu, Ximing and Dziri, Nouha and Sclar, Melanie and Chandu, Khyathi and Bhagavatula, Chandra and Choi, Yejin},
  journal={arXiv preprint arXiv:2312.01552},
  year={2023}
}

@article{clark2018think,
  title={Think you have solved question answering? Try {ARC}, the {AI2} reasoning challenge},
  author={Clark, Peter and Cowhey, Isaac and Etzioni, Oren and Khot, Tushar and Sabharwal, Ashish and Schoenick, Carissa and Tafjord, Oyvind},
  journal={arXiv preprint arXiv:1803.05457},
  year={2018}
}

@article{liu2020logiqa,
  title={{LogiQA}: A challenge dataset for machine reading comprehension with logical reasoning},
  author={Liu, Jian and Cui, Leyang and Liu, Hanmeng and Huang, Dandan and Wang, Yile and Zhang, Yue},
  journal={arXiv preprint arXiv:2007.08124},
  year={2020}
}

@inproceedings{bisk2020piqa,
  title={{PIQA}: Reasoning about physical commonsense in natural language},
  author={Bisk, Yonatan and Zellers, Rowan and Gao, Jianfeng and Choi, Yejin and others},
  booktitle={Proceedings of the AAAI Conference on Artificial Intelligence},
  volume={34},
  number={05},
  pages={7432--7439},
  year={2020}
}

@article{welbl2017crowdsourcing,
  title={Crowdsourcing multiple choice science questions},
  author={Welbl, Johannes and Liu, Nelson F and Gardner, Matt},
  journal={arXiv preprint arXiv:1707.06209},
  year={2017}
}

@article{hendrycks2020measuring,
  title={Measuring massive multitask language understanding},
  author={Hendrycks, Dan and Burns, Collin and Basart, Steven and Zou, Andy and Mazeika, Mantas and Song, Dawn and Steinhardt, Jacob},
  journal={arXiv preprint arXiv:2009.03300},
  year={2020}
}

@article{sakaguchi2021winogrande,
  title={{WinoGrande}: An adversarial winograd schema challenge at scale},
  author={Sakaguchi, Keisuke and Bras, Ronan Le and Bhagavatula, Chandra and Choi, Yejin},
  journal={Communications of the ACM},
  volume={64},
  number={9},
  pages={99--106},
  year={2021},
  publisher={ACM New York, NY, USA}
}

@article{zellers2019hellaswag,
  title={Hellaswag: Can a machine really finish your sentence?},
  author={Zellers, Rowan and Holtzman, Ari and Bisk, Yonatan and Farhadi, Ali and Choi, Yejin},
  journal={arXiv preprint arXiv:1905.07830},
  year={2019}
}

@misc{eval-harness,
  author       = {Gao, Leo and Tow, Jonathan and Abbasi, Baber and Biderman, Stella and Black, Sid and DiPofi, Anthony and Foster, Charles and Golding, Laurence and Hsu, Jeffrey and Le Noac'h, Alain and Li, Haonan and McDonell, Kyle and Muennighoff, Niklas and Ociepa, Chris and Phang, Jason and Reynolds, Laria and Schoelkopf, Hailey and Skowron, Aviya and Sutawika, Lintang and Tang, Eric and Thite, Anish and Wang, Ben and Wang, Kevin and Zou, Andy},
  title        = {The Language Model Evaluation Harness},
  month        = 07,
  year         = 2024,
  publisher    = {Zenodo},
  version      = {v0.4.3},
  doi          = {10.5281/zenodo.12608602},
  url          = {https://zenodo.org/records/12608602}
}

@incollection{gholami2022survey,
  title={A survey of quantization methods for efficient neural network inference},
  author={Gholami, Amir and Kim, Sehoon and Dong, Zhen and Yao, Zhewei and Mahoney, Michael W and Keutzer, Kurt},
  booktitle={{Low-power Computer Vision}},
  pages={291--326},
  year={2022},
  publisher={Chapman and Hall/CRC}
}

@inproceedings{jin2024comprehensive,
  title={A comprehensive evaluation of quantization strategies for large language models},
  author={Jin, Renren and Du, Jiangcun and Huang, Wuwei and Liu, Wei and Luan, Jian and Wang, Bin and Xiong, Deyi},
  booktitle={Findings of the Association for Computational Linguistics ACL 2024},
  pages={12186--12215},
  year={2024}
}

@article{zhu2024survey,
  title={A Survey on Model Compression for Large Language Models},
  author={Zhu, Xunyu and Li, Jian and Liu, Yong and Ma, Can and Wang, Weiping},
  journal={Transactions of the Association for Computational Linguistics},
  volume={12},
  pages={1556--1577},
  year={2024}
}

@inproceedings{xu2023survey,
  title={A survey on model compression and acceleration for pretrained language models},
  author={Xu, Canwen and McAuley, Julian},
  booktitle={Proceedings of the AAAI Conference on Artificial Intelligence},
  volume={37},
  number={9},
  pages={10566--10575},
  year={2023}
}

@article{xu2024kdsurvey,
  title={A survey on knowledge distillation of large language models},
  author={Xu, Xiaohan and Li, Ming and Tao, Chongyang and Shen, Tao and Cheng, Reynold and Li, Jinyang and Xu, Can and Tao, Dacheng and Zhou, Tianyi},
  journal={arXiv preprint arXiv:2402.13116},
  year={2024}
}

@article{gou2021knowledge,
  title={Knowledge distillation: A survey},
  author={Gou, Jianping and Yu, Baosheng and Maybank, Stephen J and Tao, Dacheng},
  journal={International Journal of Computer Vision},
  volume={129},
  number={6},
  pages={1789--1819},
  year={2021},
  publisher={Springer}
}

@inproceedings{wang2020structured,
  title={Structured Pruning of Large Language Models},
  author={Wang, Ziheng and Wohlwend, Jeremy and Lei, Tao},
  booktitle={Proceedings of the 2020 Conference on Empirical Methods in Natural Language Processing (EMNLP)},
  pages={6151--6162},
  year={2020}
}

@article{yang2024survey,
  title={Survey on knowledge distillation for large language models: methods, evaluation, and application},
  author={Yang, Chuanpeng and Zhu, Yao and Lu, Wang and Wang, Yidong and Chen, Qian and Gao, Chenlong and Yan, Bingjie and Chen, Yiqiang},
  journal={ACM Transactions on Intelligent Systems and Technology},
  publisher={ACM New York, NY}
}

@article{cheng2024survey,
  title={A survey on deep neural network pruning: Taxonomy, comparison, analysis, and recommendations},
  author={Cheng, Hongrong and Zhang, Miao and Shi, Javen Qinfeng},
  journal={IEEE Transactions on Pattern Analysis and Machine Intelligence},
  volume = 46,
  number = 12,
  pages = {10558--10578},
  year={2024},
  publisher={IEEE}
}

@article{stanton2021does,
  title={Does knowledge distillation really work?},
  author={Stanton, Samuel and Izmailov, Pavel and Kirichenko, Polina and Alemi, Alexander A. and Wilson, Andrew G.},
  journal={Advances in Neural Information Processing Systems},
  volume={34},
  pages={6906--6919},
  year={2021}
}

@article{gu2023minillm,
  title={{MiniLLM}: Knowledge distillation of large language models},
  author={Gu, Yuxian and Dong, Li and Wei, Furu and Huang, Minlie},
  journal={arXiv preprint arXiv:2306.08543},
  year={2023}
}

@article{kim2019qkd,
  title={{QKD}: Quantization-aware knowledge distillation},
  author={Kim, Jangho and Bhalgat, Yash and Lee, Jinwon and Patel, Chirag and Kwak, Nojun},
  journal={arXiv preprint arXiv:1911.12491},
  year={2019}
}

@article{liu2023llm,
  title={{LLM-QAT}: Data-free quantization aware training for large language models},
  author={Liu, Zechun and Oguz, Barlas and Zhao, Changsheng and Chang, Ernie and Stock, Pierre and Mehdad, Yashar and Shi, Yangyang and Krishnamoorthi, Raghuraman and Chandra, Vikas},
  journal={arXiv preprint arXiv:2305.17888},
  year={2023}
}

@inproceedings{whittaker2001quantization,
  title={Quantization-based language model compression.},
  author={Whittaker, Edward WD and Raj, Bhiksha},
  booktitle={INTERSPEECH},
  pages={33--36},
  year={2001}
}

@article{muralidharan2024compact,
  title={Compact language models via pruning and knowledge distillation},
  author={Muralidharan, Saurav and Turuvekere Sreenivas, Sharath and Joshi, Raviraj and Chochowski, Marcin and Patwary, Mostofa and Shoeybi, Mohammad and Catanzaro, Bryan and Kautz, Jan and Molchanov, Pavlo},
  journal={Advances in Neural Information Processing Systems},
  volume={37},
  pages={41076--41102},
  year={2024}
}

@article{kim2021pqk,
  title={{PQK}: model compression via pruning, quantization, and knowledge distillation},
  author={Kim, Jangho and Chang, Simyung and Kwak, Nojun},
  journal={arXiv preprint arXiv:2106.14681},
  year={2021}
}

@article{radford2019language,
  title={Language Models are Unsupervised Multitask Learners},
  author={Radford, Alec and Wu, Jeff and Child, Rewon and Luan, David and Amodei, Dario and Sutskever, Ilya},
  journal = {OpenAI Blog},
  volume = 1,
  number = 8,
  page = 9,
  year={2019}
}

@article{brown2020language,
  title={Language models are few-shot learners},
  author={Brown, Tom and others},
  journal={Advances in Neural Information Processing Systems},
  volume={33},
  pages={1877--1901},
  year={2020}
}

@article{chang2024survey,
  title={A survey on evaluation of large language models},
  author={Chang, Yupeng and others},
  journal={ACM Transactions on Intelligent Systems and Technology},
  volume={15},
  number={3},
  pages={1--45},
  year={2024},
  publisher={ACM New York, NY}
}

@misc{qwen3technicalreport,
      title={Qwen3 Technical Report}, 
      author={Qwen Team},
      year={2025},
      eprint={2505.09388},
      archivePrefix={arXiv},
      primaryClass={cs.CL},
      url={https://arxiv.org/abs/2505.09388}, 
}

@inproceedings{kwon2023efficient,
  title={Efficient Memory Management for Large Language Model Serving with PagedAttention},
  author={Woosuk Kwon and Zhuohan Li and Siyuan Zhuang and Ying Sheng and Lianmin Zheng and Cody Hao Yu and Joseph E. Gonzalez and Hao Zhang and Ion Stoica},
  booktitle={Proceedings of the ACM SIGOPS 29th Symposium on Operating Systems Principles},
  year={2023}
}

@article{li2026siamesenorm,
  title={SiameseNorm: Breaking the Barrier to Reconciling Pre/Post-Norm},
  author={Li, Tianyu and Han, Dongchen and Cao, Zixuan and Huang, Haofeng and Zhou, Mengyu and Chen, Ming and Zhao, Erchao and Jiang, Xiaoxi and Jiang, Guanjun and Huang, Gao},
  journal={arXiv preprint arXiv:2602.08064},
  year={2026}
}

@inproceedings{
gromov2025the,
title={The Unreasonable Ineffectiveness of the Deeper Layers},
author={Andrey Gromov and Kushal Tirumala and Hassan Shapourian and Paolo Glorioso and Dan Roberts},
booktitle={The Thirteenth International Conference on Learning Representations},
year={2025},
url={https://openreview.net/forum?id=ngmEcEer8a}
}

@inproceedings{
su2026gptailor,
title={{GPT}ailor: Large Language Model Pruning Through Layer Cutting and Stitching},
author={Guinan Su and Li Shen and Lu Yin and Shiwei Liu and Yanwu Yang and Jonas Geiping},
booktitle={The Fourteenth International Conference on Learning Representations},
year={2026},
url={https://openreview.net/forum?id=yCTpYe3UOL}
}

@inproceedings{
wang2025layer,
title={Layer as Puzzle Pieces: Compressing Large Language Models through Layer Concatenation},
author={Fei Wang and Li Shen and Liang Ding and Chao Xue and Ye Liu and Changxing Ding},
booktitle={The Thirty-ninth Annual Conference on Neural Information Processing Systems},
year={2025},
url={https://openreview.net/forum?id=enhFXzKii4}
}

@inproceedings{
li2025trthepruner,
title={T\'yr-the-Pruner: Structural Pruning {LLM}s via Global Sparsity Distribution Optimization},
author={Guanchen Li and Yixing Xu and Zeping Li and Ji Liu and Xuanwu Yin and Dong Li and Emad Barsoum},
booktitle={The Thirty-ninth Annual Conference on Neural Information Processing Systems},
year={2025},
url={https://openreview.net/forum?id=rAuRLePL2R}
}

@inproceedings{wolf-etal-2020-transformers,
    title = "Transformers: State-of-the-Art Natural Language Processing",
    author = "Thomas Wolf and Lysandre Debut and Victor Sanh and Julien Chaumond and Clement Delangue and Anthony Moi and Pierric Cistac and Tim Rault and Rémi Louf and Morgan Funtowicz and Joe Davison and Sam Shleifer and Patrick von Platen and Clara Ma and Yacine Jernite and Julien Plu and Canwen Xu and Teven Le Scao and Sylvain Gugger and Mariama Drame and Quentin Lhoest and Alexander M. Rush",
    booktitle = "Proceedings of the 2020 Conference on Empirical Methods in Natural Language Processing: System Demonstrations",
    month = oct,
    year = "2020",
    address = "Online",
    publisher = "Association for Computational Linguistics",
    url = "https://www.aclweb.org/anthology/2020.emnlp-demos.6",
    pages = "38--45"
}
\onecolumn
\newpage
\appendix

\section{LaCo Implementation}
\label{app:laco_implementation}

\citet{yang-etal-2024-laco} define LaCo through a layer-collapse operation based on what they refer to as \textit{Reserving-Differences-while-Seeking-Common}. For layer $l$, with parameters $\theta_l$, and the following $m$ consecutive layers, the paper defines the merged parameters as follows:

\begin{equation}
\theta_l^{*}
=
\theta_l + (\theta_{l+1}-\theta_l)+\cdots+(\theta_{l+m}-\theta_l)
=
\theta_l+\sum_{k=1}^{m}(\theta_{l+k}-\theta_l)
\end{equation}

The merged layer is thus intended to accumulate the layer-wise parameter differences of the collapsed span into the surviving layer, after which layers $l+1,\ldots,l+m$ are discarded.

However, the version released in the official LaCo implementation, available at \url{https://github.com/yangyifei729/LaCo}, differs from this formulation. In the released compression procedure, the same weights are updated repeatedly, so that each new update overwrites the result of the previous one rather than accumulating all intermediate layer differences. Consequently, the implemented procedure does not produce the cumulative weight-difference merge defined in the paper, and is therefore equivalent to layer dropping on the updated parameter subset. 

This can be shown algebraically by expanding the successive updates applied iteratively during the compression of a span from layer $l$ to layer $l+m$. Let $\theta_l^*$ denote the collapsed layer produced by the implementation. Starting from the base layer parameters $\theta_l$, the released procedure updates the same tensors iteratively across the span. After the first update, using layer $l+1$, one obtains:
\begin{equation}
\theta_l^* = \theta_l + (\theta_{l+1} - \theta_l) = \theta_{l+1}
\end{equation}
The next update, applied to the already modified collapsed layer using layer $l+2$, yields
\begin{equation}
\theta_l^* = \theta_{l+1} + (\theta_{l+2} - \theta_{l+1}) = \theta_{l+2}
\end{equation}
Each subsequent update follows the same pattern, so that the current collapsed layer is replaced by the parameters of the next layer in the span. Continuing this iterative process until the final update at layer $l+m$ yields:
\begin{equation}
\theta_l^* = \theta_{l+m-1} + (\theta_{l+m} - \theta_{l+m-1}) = \theta_{l+m}
\end{equation}

Hence, for the parameters actually modified by the released implementation, the final collapsed layer is not the cumulative merge defined in the LaCo paper, but actually the last layer in the collapsed span:
\begin{equation}
\theta_l^* = \theta_{l+m}
\end{equation}

Once layers $l+1,\ldots,l+m$ are removed, the implemented procedure is therefore algebraically equivalent to layer dropping, except for the last layer in the compressed span.







\section{GeLaCo Solution Decoding}
\label{app:decoding_algorithm}

Given a solution
\[
x = \bigl(
t,\lambda^{\mathrm{att}},\lambda^{\mathrm{ffn}},
\hat{\lambda}^{\mathrm{att}},\hat{\lambda}^{\mathrm{ffn}},
g^{\mathrm{att}},g^{\mathrm{ffn}}
\bigr)
\]
we decode it into an ordered sequence of merge operations
\(\mathcal{M} = (\mu_1,\dots,\mu_n)\). Each operation
\[
\mu_i =
\bigl(
\mathrm{base}_i,\,
\mathrm{end}_i,\,
\lambda^{\mathrm{att}_i},\,
\lambda^{\mathrm{ffn}_i},\,
g_i^{\mathrm{att}},\,
g_i^{\mathrm{ffn}}
\bigr)
\]
collapses a contiguous span of layers from \(\mathrm{base}_i\) to \(\mathrm{end}_i\) into a single layer, using per-layer attention and feed-forward coefficients
\(\lambda^{\mathrm{att}}_i\) and
\(\lambda^{\mathrm{ffn}}_i\), together with one residual gate for each branch. Applying \(\mathcal{M}\) yields the compressed model.

Decoding starts from one segment per original layer and repeatedly merges adjacent segments until the target depth is reached. The vector \(t\) assigns a priority to each boundary between consecutive layers and determines when that boundary becomes active. Each original layer is assigned one attention coefficient and one feed-forward coefficient, \((\lambda_i^{\mathrm{att}},\lambda_i^{\mathrm{ffn}})\). When the \(r\)-th merge creates a new segment, that segment receives \((\hat{\lambda}_r^{\mathrm{att}},\hat{\lambda}_r^{\mathrm{ffn}})\), and the merge uses gates \((g_r^{\mathrm{att}},g_r^{\mathrm{ffn}})\). The decoder also returns an alias map \(\mathrm{alias}\), where \(\mathrm{alias}(i)\) is the last original layer represented by compressed layer \(i\); this map is used during fitness evaluation.

\clearpage
\begin{algorithm}[H]
\caption{GeLaCo solution decoding}
\label{algo:decoding_algorithm}

\begin{algorithmic}[1]
\Require solution \(x\), target depth \(T\), maximum number of merges \(R\)
\Ensure merge sequence \(\mathcal{M}\) and alias map \(\mathrm{alias}\)

\State Extract boundary priorities \(t\), initial coefficients
\((\lambda^{\mathrm{att}},\lambda^{\mathrm{ffn}})\), new-segment coefficients
\((\hat{\lambda}^{\mathrm{att}},\hat{\lambda}^{\mathrm{ffn}})\), and gates
\((g^{\mathrm{att}},g^{\mathrm{ffn}})\) from \(x\)
\State Initialize \(S \gets (S_0,\dots,S_{L-1})\) with \(S_i=[i,i]\)
\State Assign \((\lambda_i^{\mathrm{att}},\lambda_i^{\mathrm{ffn}})\) to segment \(S_i\)

\State \(A \gets \emptyset\), \(\mathcal{M} \gets [\,]\), \(r \gets 0\)

\For{each priority level \(\tau\) among valid boundaries, in decreasing order}
    \If{\(|S| \le T\) or \(r = R\)}
        \State \textbf{break}
    \EndIf

    \State Mark as active every valid boundary whose priority is \(\tau\)

    \While{\(|S| > T\) and \(r < R\)}
        \State \(q \gets |S| - T\)
        \State Initialize the candidate list as empty \(\mathcal{Q} \gets \emptyset\)
        \State Partition \(S\) into the largest possible contiguous groups such that
        \Statex \hspace{\algorithmicindent} every boundary between consecutive segments in the group is active

        \For{each group spanning positions \(l,\dots,u\)}
            \If{\(u=l\)}
                \State \textbf{continue}
            \EndIf

            \State Let \(p_{min}\) be the smallest priority among the boundaries inside the group
            \If{\(p_{min} \ne \tau\)}
                \State \textbf{continue}
            \EndIf

            \If{\(u-l > q\)}
                \State Trim the group from the left by setting \(l \gets u-q\)
            \EndIf

            \State Add the candidate group \(l,\dots,u\) to the candidate list \(\mathcal{Q}\)
        \EndFor

        \If{the candidate list \(\mathcal{Q}\) is empty}
            \State \textbf{break}
        \EndIf

        \State Choose the candidate group \(m,\dots,n\) that would remove the largest
        \Statex \hspace{\algorithmicindent} number of segments if merged; breaking ties by later ending position \(n\), then later starting position \(m\)

        \State \(\boldsymbol{\lambda}^{\mathrm{att}}_r \gets\) attention coefficients of segments \(S_m,\dots,S_n\)
        \State \(\boldsymbol{\lambda}^{\mathrm{ffn}}_r \gets\) feed-forward coefficients of segments \(S_m,\dots,S_n\)

        \State Append $
        \mu_r \gets
        \bigl(
        m,\,
        n,\,
        \boldsymbol{\lambda}^{\mathrm{att}}_r,\,
        \boldsymbol{\lambda}^{\mathrm{ffn}}_r,\,
        g_r^{\mathrm{att}},\,
        g_r^{\mathrm{ffn}}
        \bigr)$
        to \(\mathcal{M}\)

        \State Replace \(S_m,\dots,S_n\) by one segment covering the same original layer span
        \State Assign \((\hat{\lambda}_r^{\mathrm{att}},\hat{\lambda}_r^{\mathrm{ffn}})\) to the new segment
        \State \(r \gets r + 1\)
    \EndWhile
\EndFor

\For{\(i = 0,\dots,|S|-1\)}
    \State \(\mathrm{alias}(i) \gets\) the last original layer covered by \(S_i\)
\EndFor

\State \Return \(\mathcal{M}, \mathrm{alias}\)
\end{algorithmic}
\end{algorithm}
\clearpage

\section{Search Optimization}
\label{app:cache}

\begin{figure}[h!]
    \centering
    \includegraphics[width=0.5\linewidth]{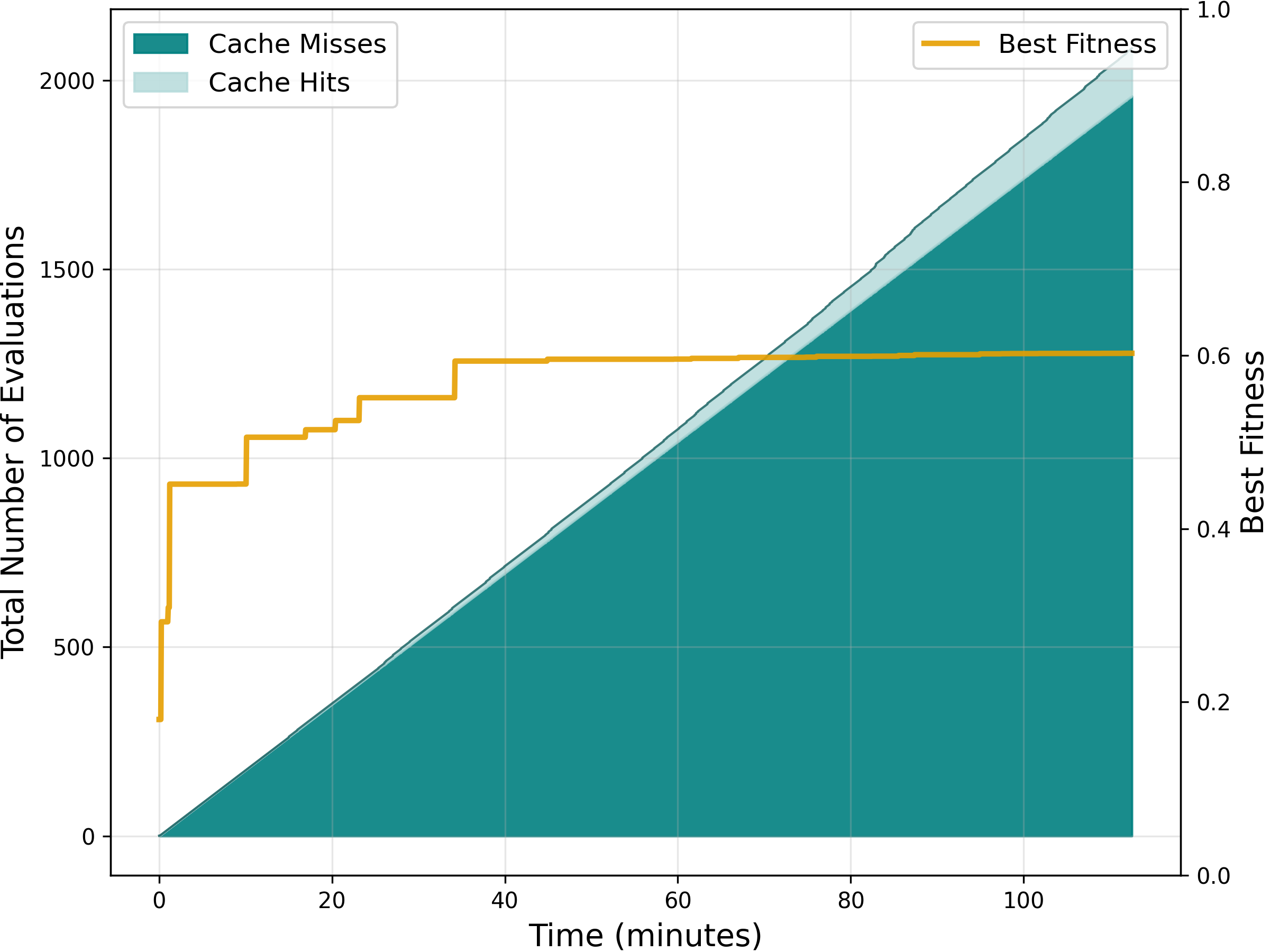}
    \caption{Processing time over number of fitness evaluations and cached solutions, for a Llama-3.1 8B model trained on a single A40 GPU with 48GB vRAM.}
    \label{fig:efficiency}
\end{figure}

GeLaCo search is optimized for fast solution evaluation. This is achieved by cloning the model and applying the compression operations directly on device. By avoiding repeated model reloading and minimizing the overhead of constructing each candidate, the average evaluation time is 3.44 seconds per candidate. We also make use of a cache to avoid recomputing the fitness of previously evaluated candidates. Since the solutions include continuous parameters, cache matching is performed after rounding floating-point values to three decimal places. Figure \ref{fig:efficiency} shows the cumulative number of evaluations together with cache hits and cache misses over time for a Llama-3.1 8B run on a single A40 GPU with 48GB of VRAM, taking approximately 2 hours to complete the search.

As the search progresses, the number of cached solutions increases, leading to a reduction in processing time as the search becomes more exploitative and the population of candidates becomes more uniform, as shown in Figure~\ref{fig:efficiency}. The number of cache hits could be increased via a reduction in precision for the continuous parameters of the solution.

\section{Post-training Hyper-parameters}
\label{app:hyperparms}

The following configuration summarizes the main parameters used for post-training the foundational Llama-3.1-8B model.
\begin{tcolorbox}[title={Llama 3.1 8B Hyper-parameters},
    enhanced, 
    breakable,
    skin first=enhanced,
    skin middle=enhanced,
    skin last=enhanced,
]
\begin{verbatim}
--dataset_name HuggingFaceFW/fineweb-edu
--learning_rate 2.0e-5
--num_train_epochs 1
--per_device_train_batch_size 8
--gradient_accumulation_steps 4
--bf16 True
--packing
--optim AdamW
--warmup_ratio 0.03
--weight_decay 0.0
--seed 42
\end{verbatim}
\end{tcolorbox}

\noindent The following configuration summarizes the main post-training parameters of the instruction-tuned Llama 3.1 8B Instruct model.

\begin{tcolorbox}[title={Llama 3.1 8B Instruct Hyperparameters},
    enhanced, 
    breakable,
    skin first=enhanced,
    skin middle=enhanced,
    skin last=enhanced,
]
\begin{verbatim}
--dataset_name MBZUAI/LaMini-instruction
--learning_rate 2.0e-5
--num_train_epochs 2
--per_device_train_batch_size 8
--gradient_accumulation_steps 4
--bf16 True
--packing
--optim AdamW
--warmup_ratio 0.03
--weight_decay 0.0
--seed 42
\end{verbatim}
\end{tcolorbox}

\section{Throughput Analysis}
\label{app:throughput}
Table~\ref{tab:llama31_throughput} reports the inference throughput (in tokens/s) for Llama-3.1 8B as a function of batch size across several compression ratios (0.125, 0.187, and 0.312). In the standard Transformers inference pipeline \cite{wolf-etal-2020-transformers}, GeLaCo and EvoPress exhibit near-identical throughput at matched compression ratios, which is expected since both approaches reduce compute primarily by removing entire Transformer blocks. At batch size 256 and compression ratio 0.187, the throughput increases from roughly 1.3k tokens/s for the dense model to $\sim$1.6k tokens/s. In contrast, LLM-Pruner yields smaller speedups ($\sim$ 1.4k tokens/s), indicating that channel pruning does not translate as directly into end-to-end speed gains compared to reducing network depth. 

\begin{table}[h!]
\centering
\resizebox{0.6\columnwidth}{!}{\begin{tabular}{lccccccc}
\toprule
\textbf{Method} & \textbf{Ratio} &
\textbf{BS=1} & \textbf{BS=8} & \textbf{BS=32} &
\textbf{BS=128} & \textbf{BS=256} & \textbf{BS=512} \\
\midrule
Dense & -- 
& 39.8 & 269.6 & 719.7 & 1168.5 & 1266.5 & -- \\

\midrule
GeLaCo & 0.125 
& 44.9 & 304.8 & 816.4 & 1328.4 & 1444.5 & -- \\
GeLaCo & 0.187 
& 47.9 & 325.3 & 874.3 & 1424.7 & 1558.3 & -- \\
GeLaCo & 0.312 
& 55.2 & 377.1 & 1019.0 & 1670.6 & 1823.4 & -- \\

\midrule
EvoPress & 0.125 
& 44.7 & 303.7 & 814.0 & 1327.4 & 1443.0 & -- \\
EvoPress & 0.187 
& 47.6 & 322.3 & 869.1 & 1420.3 & 1550.0 & -- \\
EvoPress & 0.312 
& 54.9 & 374.8 & 1012.5 & 1664.6 & 1825.0 & -- \\

\midrule
LLM-Pruner & 0.125 
& 43.8 & 291.2 & 797.5 & 1299.7 & 1416.4 & -- \\
LLM-Pruner & 0.187 
& 45.8 & 305.9 & 815.4 & 1312.0 & 1417.8 & -- \\
LLM-Pruner & 0.312 
& 51.5 & 351.2 & 1015.7 & 1642.4 & 1801.4 & -- \\

\midrule
Dense (vLLM) & -- 
& 44.8 & 333.9 & 1180.0 & 3517.9 & 4857.3 & 5767.0 \\

GeLaCo (vLLM) & 0.125 
& 50.4 & 378.0 & 1330.9 & 3998.9 & 6025.2 & 6462.5 \\
GeLaCo (vLLM) & 0.187 
& 50.4 & 378.1 & 1334.9 & 3981.9 & 6035.7 & 6582.4 \\
GeLaCo (vLLM) & 0.312 
& 50.3 & 377.3 & 1333.0 & 4007.2 & 6043.8 & 6568.4 \\
\bottomrule
\end{tabular}}
\caption{Decoding throughput (tokens/s) for Llama-3.1-8B across batch sizes (BS) under the standard Transformers pipeline and vLLM. EvoPress and LLM-Pruner are omitted in the vLLM setting due to lack of support; missing entries indicate out-of-memory errors.}
\label{tab:llama31_throughput}
\end{table}

To assess the practical benefits of compression under realistic deployment conditions, we further evaluated inference throughput using vLLM \cite{kwon2023efficient}, an optimized runtime widely adopted for serving large language models. Since GeLaCo preserves the standard Transformer block structure, it is fully compatible with vLLM out-of-the-box and achieves substantial throughput gains at larger batch sizes. At batch size 512 and compression ratio 0.187, the throughput increases from $\sim$ 5.8k tokens/s for the dense checkpoint to $\sim$6.6k vtokens/s for GeLaCo, corresponding to an approx. 4$\times$ increase relative to the standard Transformers pipeline. By contrast, EvoPress checkpoints were not supported by vLLM due to architectural modifications introduced by their compression procedure, namely the selective removal of either attention or feed-forward components within a Transformer block, while LLM-Pruner alters the channel dimensions within each block, preventing direct execution under vLLM. These differences highlight a practical deployment advantage of GeLaCo.

\section{Evaluation Rubric}
\label{app:eval-rubric}
The rubric used for the Just-Eval evaluation with Prometheus 2 as the judge model is provided below.
\begin{tcolorbox}[title={Just-Eval Evaluation Rubric},
    enhanced, 
    breakable,
    skin first=enhanced,
    skin middle=enhanced,
    skin last=enhanced,
]
\scriptsize
\begin{verbatim}
{
    "criteria": "Does the model provide a high-quality response that accurately addresses
    the user's question or request?",
    "score1_description": "The response is completely off-topic, nonsensical, factually
    incorrect in fundamental ways, or potentially harmful with no redeeming value.",
    "score2_description": "The response misses the main point or contains significant
    errors, but shows some understanding of the topic and provides limited value to the
    user.",
    "score3_description": "The response addresses the user's question with basic accuracy 
    and relevance, though it may lack depth, have some minor errors, or miss important
    nuances.",
    "score4_description": "The response is good quality with strong accuracy, relevance,
    clarity, and helpfulness, containing only minor areas for improvement.",
    "score5_description": "The response is excellent quality, demonstrating high accuracy,
    perfect relevance, exceptional clarity, and outstanding helpfulness that fully meets
    or exceeds user expectations."
}
\end{verbatim}
\end{tcolorbox}
\end{document}